\newif\ifdraft\drafttrue
\documentclass{article}

\usepackage{fullpage}
\usepackage[square, numbers, sort]{natbib}

\usepackage[utf8]{inputenc} 
\usepackage[T1]{fontenc}    
\usepackage{hyperref}       
\usepackage{booktabs}       
\usepackage{longtable}
\usepackage{amsfonts}       
\usepackage{nicefrac}       
\usepackage{microtype}      

\usepackage{mathtools}
\usepackage{bbm, bm}

\usepackage{algorithmic} 
\usepackage[ruled]{algorithm2e}

\SetAlFnt{\small}
\SetAlCapFnt{\small}
\SetAlCapNameFnt{\small}
\SetAlCapHSkip{0pt}

\usepackage{authblk}
\usepackage{array, multirow, adjustbox}
\usepackage{arydshln}
\usepackage{blindtext}
\usepackage{bm}
\usepackage{booktabs} 
\usepackage{comment}
\usepackage{enumitem}
\usepackage{graphicx}
\usepackage{hyperref}
\usepackage{multicol}
\usepackage{subcaption}
\usepackage{tikz} 
\usepackage{url}
\usepackage{xcolor}

\newcommand{\R}{\mathbb{R}}
\newcommand{\E}{\mathbb{E}}

\DeclareMathOperator*{\argmin}{arg\,min}

\newcommand{\Prb}{\mathbb{P}} 

\newcommand{\Ind}{\mathbb{I}} 
\usepackage{subcaption}

\newtheorem{theorem}{Theorem}

\newtheorem{assumption}[theorem]{Assumption}

\newcommand{\cX}{\mathcal{X}} 
\newcommand{\cF}{\mathcal{F}} 
\newcommand{\cost}{c} 

\usepackage{xcolor}

\title{Algorithmic Recourse Under Competition}
\author{Shahin Jabbari\thanks{Drexel University, shahin@drexel.edu}}
\date{}

\begin{document}

\maketitle

\begin{abstract}
Algorithmic recourse provides individuals who have received undesirable outcomes from machine learning models with suggestions for minimum-cost improvements to achieve the desired outcome. A central assumption when computing recourse is that the decision rule remains fixed throughout the recourse implementation phase. We challenge this assumption in settings where individuals compete for limited resources. In such settings, widespread recourse implementation can change the acceptance threshold even when the scoring model that is used to evaluate individuals remains the same. This change in acceptance threshold can, in turn,  invalidate the original recourse recommendations (i.e., following the recourse may not lead to the desired outcome). To address this problem, we introduce a framework called \emph{recourse under competition} that jointly optimizes for recommendation recipients and the recommended score target they need to satisfy to balance the recourse cost and post-shift validity among initially rejected individuals. We develop an algorithm based on the Implicit Function Theorem and empirically analyze its performance. Experiments on synthetic and real datasets show that personalized score targets can achieve higher validity, albeit at a higher cost. In contrast, common score targets generally offer favorable cost-validity trade-offs for lower to medium validity values.
\end{abstract}

\section{Introduction}
\label{sec:intro}
With the rapid deployment of machine learning models in critical
domains and the major impact of their decisions on people's
livelihoods, a surge of recent work in responsible machine learning
aims to make these models fair~\citep{BerkHJKR18, BarocasHN19, ZafarVGG17, HardtPS16},
transparent~\citep{Rudin19, LakkarajuBL16}, and
explainable~\citep{RibeiroSG16, LundbergL17, SmilkovTKVW17}.
A recent line of work within the explainability literature,
termed \emph{algorithmic recourse}~\citep{WachterMR18, UstunSL19},
involves the decision-maker providing individuals who received
an undesirable label (e.g., one whose loan request was denied)
with minimum-cost improvement suggestions to achieve the
desired label.

A central assumption when computing recourse is that the decision rule
remains fixed during the time needed for individuals to
implement their modifications. However, this assumption can fail
in settings where individuals compete for limited resources, and only a
limited number or fraction of individuals can be accepted
(e.g., a fixed number of approved loans or university admissions).
As individuals implement recourse, they alter the population's
distribution. To maintain the acceptance rate dictated by limited resources,
the decision-maker recomputes the acceptance threshold for the
updated population, even when its scoring model remains unchanged.
The updated threshold can invalidate the initially prescribed recourse, i.e.,
following the suggestions may no longer lead to the desired
outcome. Prior work has shown
reaching the original acceptance threshold is insufficient to
ensure acceptance under competition~\citep{FonsecaBABS23}. This motivates designing
recourse recommendation strategies that explicitly anticipate the
competition induced by their implementation.

In this work, we propose \emph{recourse under competition},
a framework that explicitly anticipates future updates in 
acceptance threshold in competitive environments. In this formulation,
the individuals are restricted by a budget for their feature modification, and
the decision-maker simultaneously should decide which individuals should be instructed 
to implement recourse and what score target each such individual should aim for. 
Our main goal is to understand the trade-off between the
expected recourse cost across the population of rejected individuals against the validity of recourse, 
after accounting for distribution changes and acceptance threshold updates 
caused by recourse implementation. 

\paragraph{Our Results and Contributions.}
We formalize \emph{recourse under competition}, a
framework for balancing recourse cost and post-shift validity through
population-level recommendation policies. We develop an optimization algorithm based on the Implicit Function Theorem
that anticipates future updates in the selection threshold when recommending recourse.

We evaluate our algorithm on four real datasets and
two synthetic datasets to study the trade-off 
between the cost of recourse implementation and its validity after competitive threshold updates. 
We observe empirical convergence of the smooth training objective across 
different datasets and model classes. 
Our main findings indicate that anticipating recourse-induced threshold changes
can significantly improve the validity of recourse compared to the original-threshold baseline
that do not consider competition when providing recourse. We also find that
personalized score targets can achieve higher validity at greater
cost, while common score targets often offer better tradeoffs
at lower to medium validity values.
Sensitivity experiments show that
larger budgets and admission rates generally increase
validity. We also study the effect of these selected policies on the
distribution of recourse cost and targets across populations and their disparities across 
subpopulations.
\section{Related Work}
\label{sec:related}

\paragraph{Algorithmic Recourse.}
Recourse is a post-hoc counterfactual explanation that aims to provide the lowest-cost modification that changes the prediction for a given input with an undesirable predicted label under the current model ~\cite{WachterMR18, UstunSL19}. Since its introduction, different formulations have been used to model the optimization problem in recourse~\cite{LooverenK21, RawalL20, SlackHLS21, PawelczykDHKL23, GargNS25, KarimiBBV20, KarimiKSV20, KanamoriTKI24, BewleyAM+24, VermaDH22, ChenEV+25} or study additional aspects such as fairness~\cite{GuldoganZS+23, GuptaNR+19, HeidariNG19, GaoL23, BoxerN25, PerelloCZ+25}, repeated dynamics~\cite{FonsecaBABS23,BellFA+24,EhyaeiSS25}, improvement~\cite{KonigFG23, AvasaralaGJ+26, KonigFF+25}, effects of temporal data~\cite{BuligaDG+25} and potential harms~\cite{FokkemaGE24}. See \cite{VermaDH20, KarimiBSV23} for surveys.

\citet{WachterMR18}~and~ \citet{PawelczykDHKL23} focus on score-based classifiers, generating feature modifications that help individuals attain a target score. In contrast, ~\citet{UstunSL19} addresses binary classifiers and requires that recourse actions lead to a change in the predicted label. Conceptually, the score-based formulation can be viewed as a relaxation of the label-based setting. We also adopt a score-based formulation in our work subject to a recourse budget. Our main point of departure from all these works is that we optimize recourse at the population level by simultaneously considering the effect of recourse on all individuals. 

\paragraph{Robust Recourse.}
Although most prior work assumes a static setting, several recent works have focused on robustness under uncertainty about the data or the classifier. \citet{UpadhyayJL21} and \citet{NguyenBN+22} propose recourse mechanisms that remain effective under model shift induced by distributional shift. \citet{GuoJC+23} introduces a robust training procedure that jointly trains the classifier and a recourse model under data shifts. Similar to algorithmic recourse, many different variations and formulations of robust recourse have also been proposed (see, e.g., \cite{LeofanteP24,YetukuriHVUL24, NguyenBN23, DuttaLMTM22, CheonWF+25, KyawKJ26, KayasthaGJ26, TurbalVS25} and \cite{JiangLR+24a} for a survey). Our setting considers a different source of recourse invalidation:
recommendations change the applicant population and thereby the
capacity-constrained acceptance threshold, even though the scoring
function remains \emph{fixed}.

\paragraph{Recourse Over Time with Limited Resources.}
Several works studied the temporal dynamics of recourse by analyzing how recourse recommendations change over time in settings where individuals repeatedly interact with a decision-making system under resource constraints~\cite{FonsecaBABS23, ToniTL+24, LiuLC+25, AltmeyerABDDL23}.~\citet{SegalGYD24} modeled limited resources using a 0/1 knapsack formulation.~\citet{rldurable} formulated the long-term effects of recourse and resource limitations as a partially observable MDP. Conceptually, the closest related work models collective recourse through penalties
for population congestion~\cite{EhyaeiSS25} or through capacitated
matching between applicants and providers~\cite{KhotanlouLK26}. Direct comparison to these works is difficult due to different modeling choices.

\paragraph{Performative Prediction.}
\citet{PerdomoZMH20} introduced performative prediction, where predictions affect behavior; the data distribution can shift in response, making it hard to train a model that performs well after a distribution shift. They show that standard empirical risk minimization may fail when the model deployment changes the data generation process. They distinguish between two key notions: performative optimality, where a model minimizes empirical loss on the distribution it induces, and performative stability, where the selected model matches the model induced by the deployment. They provided algorithms to compute performatively stable points and sufficient conditions for their existence. They also show that performatively optimal points are in a close neighborhood of performatively stable points. These algorithms were later extended to compute performatively optimal solutions under specific assumptions on the input distributions~\cite{IzzoYZ21, IzzoZY22}. Building on this, \citet{Mendler-DunnerDW22} propose methods to anticipate performative shifts by learning to predict from predictions, offering stability-aware model selection strategies~\cite{KimP23, PerdomoBH+25}. Through a causality lens,~\citet{KonigFF+25} study sufficient conditions under which recourse remains valid after performativity. See~\cite{HardtM23} for a survey. Our objective shares the performative perspective of evaluating a
policy under the distribution it induces, though our models do not satisfy the strong convexity and smoothness assumptions that are commonly used in the performative prediction literature. 
\section{Problem Formulation}
\label{sec:framework}

Let $\cX\subseteq\R^d$ be the instance space representing individuals. 
Let $D$ be an \emph{unknown} population distribution. An unknown ground-truth 
function $f^*:\cX\to[0,1]$ determines the qualification of individuals. 
The decision-maker can learn a proxy for $f^*$ by performing empirical risk minimization
using a parameterized hypothesis class
$\cF=\{f_w:\cX\to[0,1]\mid w\in\mathcal W\subseteq\R^n\}$:
\begin{equation}
w_0\in\argmin_{w\in\mathcal W}
\E_{x\sim D}\!\left[
\ell_{\mathrm{erm}}\bigl(f_w(x),f^*(x)\bigr)
\right],
\label{eq:initial-erm}
\end{equation}
where $\ell_{\mathrm{erm}}$ is a convex loss function. Throughout, we refer to $f_{w_0}$
as the \emph{scoring function}.

Individuals compete
for a resource that can be allocated to an $\alpha\in(0,1)$ fraction
of the population. 
Assume the initial score distribution is continuous. The decision-maker determines the 
initial allocation by selecting the highest-scoring $\alpha$ fraction, with threshold
\begin{equation}
t_0=\inf\left\{
t\in[0,1]:
\Prb_{x\sim D}[f_{w_0}(x)\ge t]\le\alpha
\right\}
\label{eq:initial-threshold}
\end{equation}
and the decision rule
\begin{equation}
\hat y_0(x)=\Ind[f_{w_0}(x)\ge t_0].
\label{eq:initial-decision}
\end{equation}
Let $X^-=\{x:\hat y_0(x)=0\}$ and
$X^+=\{x:\hat y_0(x)=1\}$ denote the initially rejected and accepted
individuals, respectively. Algorithmic recourse recommends feature modifications intended to
improve an individual's outcome \citep{WachterMR18,UstunSL19}.
Let $\cost:\cX\times\cX\to\R_+$ denote a cost function for feature modification that satisfy $\cost(x,x)=0$, and let
$B\ge0$ be a common recourse budget.\footnote{Individual-specific budgets can
be accommodated in the same way.} 

Unlike most prior work that provides recourse at an individual level, we focus on 
recourse at the population level.
In a \emph{population-level recourse policy}, the decision-maker not only provides a target
score for each individual to achieve but also recommends whether this recommendation should be implemented or not.
More formally, the population-level recourse policy consists of two measurable functions:
\begin{equation}
q:\cX\to[t_0,1],
\qquad
\rho:\cX\to\{0,1\}.
\label{eq:target-and-recommendation}
\end{equation}
The target $q(x)$ specifies which score an individual should attain; $\rho(x)=1$ means
that recourse is recommended (to be implemented), and $\rho(x)=0$ means the recourse is not recommended.\footnote{
We assume every recommended response is implemented.
}

For $x\in X^-$ whose target is feasible within budget, define the
\emph{best response} as
\begin{align}
\operatorname{br}(x;q)
&\in\argmin_{x'\in\cX}\cost(x,x') \nonumber\\
\text{s.t.}\quad
&f_{w_0}(x')\ge q(x),
\cost(x,x')\le B.
\label{eq:recourse-policy}
\end{align}
We assume a minimizer exists whenever the constraints are feasible. 
For infeasible targets and initially
accepted applicants, we set $\operatorname{br}(x;q)=x$ and require
$\rho(x)=0$. In particular, initially rejected applicants who cannot
reach the initial target of $t_0$ within budget always receive no recommendation. 
Among the remaining rejected applicants, the decision-maker chooses $\rho(x)$,
subject to recommending only feasible targets.

The pair $(q,\rho)$ induces the \emph{recourse map} $r$
\begin{equation}
r(x)=
\begin{cases}
\operatorname{br}(x;q), & \text{if}\quad \rho(x)=1,\\
x, & \text{if}\quad \rho(x)=0.
\end{cases}
\label{eq:implemented-policy}
\end{equation}
Thus $\operatorname{br}(x;q)$ is the candidate response to the target,
whereas $r(x)$ is the individual's resulting feature vector post-recourse.

The recourse policy can impose the following situations:
\begin{enumerate}
\item \emph{A common target:} $q(x)=q_c$ for all $x\in X^-$, with
$\rho(x)=1$ exactly for rejected applicants who can reach the target $q_c$.
Only the scalar $q_c\in[t_0,1]$ is optimized.
\item \emph{Personalized targets without selection:} $q$ is optimized and $\rho(x)=1$ 
for any rejected applicant who can reach their corresponding target $q(x)$.
\item \emph{Personalized targets with selection:} both $q$ and $\rho$
are optimized as functions of applicant features. 
\end{enumerate}
In particular, setting $q(x)=t_0$ and recommending recourse to every
feasible rejected applicant recovers minimum-cost recourse to the
original threshold as studied in prior work~\citep{FonsecaBABS23}. 

Implementing recourse will change the distribution of individuals. 
Let $D_r$ denote the distribution of $r(x)$ for $x\sim D$. We assume
the (post-recourse) qualification can still be assessed through
the ground truth function $f^*$ as formally stated below.

\begin{assumption}
\label{assumpt:stable-qualification}
The qualification function $f^*$ is invariant to population composition.
Recourse may change $f^*(x)$ to $f^*(r(x))$, but does not change the
relationship between resulting features and qualification.
\end{assumption}
Given that the distribution of individuals can change after recourse, the decision-maker
should update its threshold to satisfy the capacity constraint:
\begin{equation}
t_1(r)=\inf\left\{
t\in[0,1]:
\Prb_{x'\sim D_r}[f_{w_0}(x')\le t]\ge1-\alpha
\right\}.
\label{eq:post-threshold}
\end{equation}
Note that since we assumed $f^*$ is unchanged, the decision-maker still 
utilizes $f_{w_0}$ to set the acceptance threshold.

Recourse can create positive probability mass at a target score.
When the cutoff threshold $t_1(r)$ has positive mass, define
\begin{equation}
\beta_r=
\frac{
\alpha-\Prb_{x'\sim D_r}[f_{w_0}(x')>t_1(r)]
}{
\Prb_{x'\sim D_r}[f_{w_0}(x')=t_1(r)]
},
\label{eq:threshold-tie-probability}
\end{equation}
as the fraction of individuals with score $t_1(r)$ that can be selected without 
violating the capacity constraint. 
Otherwise set $\beta_r=0$.
For a uniform random variable $U\sim\operatorname{Unif}[0,1]$, independent of the applicant and
drawn after recourse, the future allocation rule is
\begin{equation}
\hat y_1^r(x';U)
=
\Ind[f_{w_0}(x')>t_1(r)]
+
\Ind[f_{w_0}(x')=t_1(r)]\Ind[U\le\beta_r].
\label{eq:post-decision}
\end{equation}
This rule gives the population acceptance probability $\alpha$, after
randomization at the threshold $t_1(r)$.

For the recourse map $r$ induced by $(q,\rho)$, define post-shift
validity $V$ and expected cost $C$ over all initially rejected individuals as
\begin{align}
V(q,\rho)
&=
\Prb_{x\sim D,\,U}\!\left[
\hat y_1^r(r(x);U)=1
\,\middle|\,x\in X^-
\right],
\label{eq:validity-definition}\\
C(q,\rho)
&=
\E_{x\sim D}\!\left[
\cost(x,r(x))
\,\middle|\,x\in X^-
\right].
\label{eq:expected-recourse-cost}
\end{align}
These definitions include individuals receiving no recommendation and
costs incurred by those whose recourse does not lead to acceptance.
Let $\mathcal P$ be the chosen family of measurable pairs $(q,\rho)$. 
For a fixed $\lambda>0$,
the decision-maker aims to optimize the following objective:
\begin{equation}
(q^*,\rho^*)\in\argmin_{(q,\rho)\in\mathcal P}
\left\{C(q,\rho)-\lambda V(q,\rho)\right\}.
\label{eq:validity}
\end{equation}
The parameter $\lambda$ weights validity relative to cost; it does not
impose a prescribed minimum validity. We can study the trade-off between validity
and cost by varying $\lambda$ and solving the optimization problem in Equation~\eqref{eq:validity}
for each $\lambda$.
\section{Our Algorithm}
\label{sec:algorithm}

Optimizing Equation~\eqref{eq:validity} is difficult because recommendations
change the competitive threshold, and both acceptance and the binary
recommendation decision $\rho$ are discontinuous. Our main algorithmic
idea is to use smooth surrogates for $q$ and $\rho$, smooth the acceptance
threshold, and apply the Implicit Function
Theorem (IFT) to differentiate through the resulting competitive
threshold.

As is common, we assume a sample $S=\{x_1,\ldots,x_N\}$ of size $N$
drawn independently from $D$ and optimize empirical averages instead
of expectations. Throughout this section, we assume the scoring model $f_{w_0}$ is 
learned beforehand using the empirical counterpart of Equation~\eqref{eq:initial-erm}
on a labeled sample. That supervised formulation assumes
qualification labels $f^*(x)$ are available for score training. 
We discuss how we learn this function from data in more detail 
in Section~\ref{sec:experiments}.
The recourse sample $S$ requires only features.
The model $f_{w_0}$ and initial threshold $t_0$ remain fixed during
recourse optimization.

Let $k=\lfloor\alpha N\rfloor$ and use this threshold to select the acceptance threshold of $f_{w_0}$ such that 
only $k$ individuals are selected by $f_{w_0}$. Let $S^-$ denote the rejected individuals.
We assume $|S^-|>0$ and $1\le k<N$.
For each rejected applicant, we precompute
\begin{equation}
\bar q(x)=
\max_{x'\in\cX:\,\cost(x,x')\le B}f_{w_0}(x'),
\label{eq:q-bar}
\end{equation}
assuming the maximum is attained.
In words, $\bar q(x)$  is the highest score achievable by individual $x$ under $f_{w_0}$ within budget $B$.
We refer to an individual $x$ as eligible if $x\in X^-$ and
$\bar q(x)\ge t_0$, i.e., a rejected individual who can at least
reach the initial threshold. 

Let $\sigma(z)=(1+e^{-z})^{-1}$ denote the sigmoid function.
We estimate $q$ and $\rho$ using smooth surrogates of the form
\begin{equation}
\widehat q_{\theta_1}(x)
=t_0+(\bar q(x)-t_0)\sigma\bigl(g_{\theta_1}(x)\bigr),
\label{eq:policy-parameterization-1}
\end{equation}
and
\begin{equation}
\widehat\rho_{\theta_2}(x)
=\sigma\bigl(h_{\theta_2}(x)\bigr)
\label{eq:policy-parameterization-2}
\end{equation}
for eligible applicants. The functions $g_{\theta_1}$ and
$h_{\theta_2}$ are twice continuously differentiable in their
respective parameters on neighborhoods of
$\Theta_1\subseteq\R^{m_1}$ and $\Theta_2\subseteq\R^{m_2}$.
We assume each parameter set is nonempty, compact, and convex.

The estimate $\widehat q_{\theta_1}$ is feasible by construction
and can be used to compute the best response
$\operatorname{br}(x;\widehat q_{\theta_1})$
using Equation~\eqref{eq:recourse-policy}.
For ineligible applicants, including initially accepted individuals,
we set $\widehat q_{\theta_1}(x)=t_0$,
$\widehat\rho_{\theta_2}(x)=0$, and
$\operatorname{br}(x;\widehat q_{\theta_1})=x$.

\begin{algorithm}[t!]
\caption{Recourse under Competition}
\label{therecoursealgorithm}
\textbf{Input}: Sample $S$ of individuals;
fixed model $f_{w_0}$; budget $B$; selection rate $\alpha$;
initial threshold $t_0$; parametric functions $g$ and $h$;
weight $\lambda>0$; temperature $\tau>0$;
learning rate $\eta>0$; iterations $T$. \\
\textbf{Output}: Selected target function $q$ and binary
recommendation function $\rho$.
\begin{algorithmic}[1]
\STATE Randomly initialize
$\theta_\ell^{(1)}\in\Theta_\ell$ for $\ell\in\{1,2\}$.
\STATE Let $k=\lfloor\alpha N\rfloor$ and compute $S^-$.
\STATE Precompute $\bar q$ using Equation~\eqref{eq:q-bar}
and determine eligibility.
\FOR{$j=1,\ldots,T$}
    \STATE For each eligible $x_i\in S$, evaluate
    $\widehat q_{\theta_1^{(j)}}(x_i)$ and
    $\widehat\rho_{\theta_2^{(j)}}(x_i)$ using
    Equations~\eqref{eq:policy-parameterization-1}
    and~\eqref{eq:policy-parameterization-2}.
    \STATE For each eligible $x_i\in S$, compute the best response
    using Equation~\eqref{eq:recourse-policy}.
    \STATE Form $a_i(t,\theta_1^{(j)},\theta_2^{(j)})$ for all
    $i=1,\ldots,N$ using Equation~\eqref{eq:allocation-weights}.
    \STATE Solve Equation~\eqref{eq:smoothed-capacity}
    by bisection to obtain $\hat t$.
    \STATE Compute the full gradients of
    Equation~\eqref{eq:empirical-objective} at the current iterate,
    differentiating through the best responses and using the IFT-based
    threshold derivative in Equation~\eqref{eq:empirical-threshold-gradient}.
    \STATE Update both parameter vectors by a gradient step with learning rate $\lambda$:
    \begin{equation}
    \theta_\ell^{(j+1)}
    =
    \Pi_{\Theta_\ell}\!\left(
    \theta_\ell^{(j)}
    -\eta\nabla_{\theta_\ell}
    \widehat J(\theta_1^{(j)},\theta_2^{(j)})
    \right),
    \ell\in\{1,2\},
    \label{eq:policy-update}
    \end{equation}
    where $\Pi_{\Theta_\ell}$ denotes Euclidean projection.
\ENDFOR
\STATE Set
$q(x)=\widehat q_{\theta_1^{(T)}}(x)$ and
$\rho(x)=\Ind[\widehat\rho_{\theta_2^{(T)}}(x)\ge1/2]$,
\RETURN $q$ and $\rho$.
\end{algorithmic}
\end{algorithm}

We use $\widehat\rho$ to form a differentiable relaxation of
recommendation decisions. Let $\tau>0$ be a temperature parameter.
During training, we mix the smoothed acceptance outcomes of no action
and the complete best response, with respective weights
$1-\widehat\rho_{\theta_2}(x)$ and
$\widehat\rho_{\theta_2}(x)$, defined formally as:
\begin{equation}
a(t,\theta_1,\theta_2)
=\left(1-\widehat\rho_{\theta_2}(x)\right)
\sigma\!\left(\frac{f_{w_0}(x)-t}{\tau}\right)\\
+\widehat\rho_{\theta_2}(x)
\sigma\!\left(
\frac{f_{w_0}(\operatorname{br}(x;\widehat q_{\theta_1}))-t}{\tau}
\right).
\label{eq:allocation-weights}
\end{equation}
Thus, $a$ averages outcomes rather than feature changes. 
It should be treated as an approximation that our algorithm uses in training, 
not the acceptance probability
of the final binary policy we are aiming to compute. The temperature parameter $\tau$ controls how sharply training approximates
the acceptance threshold. For small $\tau$, acceptance changes sharply
near the threshold $t$. This resembles the actual decision rule, but
gradients become concentrated around the threshold. For large $\tau$,
acceptance changes gradually over a wider score range. This provides
broader gradient signals but is a less accurate approximation of
hard acceptance.

Define $\hat t=\hat t(\theta_1,\theta_2)$ as the unique real solution of
\begin{equation}
\sum_{i=1}^{N}a_i(\hat t,\theta_1,\theta_2)=k.
\label{eq:smoothed-capacity}
\end{equation}
Note that the sum includes all applicants because initially accepted individuals
continue to compete for the resource. 

Using the quantities we defined so far, we obtain the following smooth empirical
approximation of Equation~\eqref{eq:validity}:
\begin{equation}
\widehat J(\theta_1,\theta_2)
=\frac{1}{|S^-|}\sum_{i\in S^-}
\Big[
\widehat\rho_{\theta_2}(x_i)
\cost\bigl(x_i,\operatorname{br}(x_i;\widehat q_{\theta_1})\bigr)
-\lambda a_i(\hat t,\theta_1,\theta_2)
\Big],
\label{eq:empirical-objective}
\end{equation}
where $\lambda>0$ is fixed as in the formulation.

Before stating our algorithm, we make the following response regularity assumption.
\begin{assumption}
\label{assumpt:general}
For every $x_i\in S$, the maps
$\theta_1\mapsto
f_{w_0}(\operatorname{br}(x_i;\widehat q_{\theta_1}))$
and
$\theta_1\mapsto
\cost(x_i,\operatorname{br}(x_i;\widehat q_{\theta_1}))$
are twice continuously differentiable on a neighborhood of $\Theta_1$.
\end{assumption}

Algorithm~\ref{therecoursealgorithm} alternates between computing
targets and best responses, recalibrating the competitive threshold,
and updating both parameter vectors.
Updating $\theta_1$ changes targets and best responses; updating
$\theta_2$ changes recommendation weights through
Equations~\eqref{eq:policy-parameterization-1}
and~\eqref{eq:policy-parameterization-2}.
Neither output is updated independently for each applicant.
Under Assumption~\ref{assumpt:general}, the left-hand side of
Equation~\eqref{eq:smoothed-capacity} is continuously differentiable
and has a strictly negative derivative with respect to $t$.
Therefore, the IFT ensures that $\hat t$
is differentiable in the policy parameters and gives
\begin{equation}
\nabla_{\theta_\ell}\hat t
=-\frac{
\sum_{i=1}^{N}\partial_{\theta_\ell}
a_i(\hat t,\theta_1,\theta_2)
}{
\sum_{i=1}^{N}\partial_t
a_i(\hat t,\theta_1,\theta_2)
},
\qquad \ell\in\{1,2\}.
\label{eq:empirical-threshold-gradient}
\end{equation}
Here $\partial_{\theta_\ell}$ holds the threshold fixed.
The denominator is strictly negative. Full gradients of
Equation~\eqref{eq:empirical-objective} include differentiation
through the best responses and this threshold dependence. After $T$ rounds, 
the algorithm uses the
learned weights $\theta_1^{(T)}$ to form the estimate for $\widehat q$. The learned weight
$\theta_2^{(T)}$ form estimates for $\widehat \rho$ which will get converted to binary decisions
by thresholding.

\section{Experiments}
\label{sec:experiments}
In this section, we empirically evaluate our algorithm on synthetic and real datasets. We describe our datasets in Section~\ref{sec:datasets}, our implementation details in Section~\ref{sec:implementation-details}, and our findings in Section~\ref{sec:findings}.
Our code is available \href{https://github.com/shahin-jabbari/RecourseCompetition}{here}.

\subsection{Datasets}
\label{sec:datasets}
The experiments use four real datasets (Adult, German Credit,
ACSIncome, and Give Me Some Credit) and two synthetic populations. All real
datasets are publicly available, and our code provides details on where the dataset
is downloaded from.
Adult and ACSIncome predict whether income exceeds \$50,000;
ACSIncome uses the 2018 California sample~\cite{DingHMS21}.
German Credit classifies credit risk, and Give Me Some Credit predicts
serious delinquency within two years. Favorable labels correspond to
high income, good credit, and no serious delinquency, respectively.
We generate two synthetic populations.
Each synthetic population contains $6{,}000$ three-dimensional Gaussian
observations with pairwise correlation $0.5$ and a known nonlinear
qualification function. Synthetic Nonlinear uses a sinusoidal
interaction in the logit, while Synthetic Curved uses a concave quadratic
logit. 

Preprocessing replaces missing numerical values with the corresponding
feature median and missing categorical values with their most frequent
category. Numerical features
are standardized to zero mean and unit variance, and categorical
features are one-hot encoded. Synthetic features retain their generated scale.
For each dataset, we divide the features into two sets: mutable features and immutable features.
Immutable features remain inputs to the scoring function and recourse policy, but their
encoded coordinates cannot change during recourse generation. For the synthetic dataset, we designated the first coordinate to be immutable
in both populations. Table~\ref{tab:datasets} provides a summary of datasets and the number of features 
before and after preprocessing.

\begin{table}[ht]
\centering
\begin{tabular}{lrrr}
\toprule
\textbf{Dataset} & \textbf{Size} & \textbf{Raw features} & \textbf{Processed features} \\
\midrule
Adult & 48,842 & 14 & 104 \\
German Credit & 1,000 & 20 & 61 \\
Give Me Some Credit & 150,000 & 10 & 10 \\
ACSIncome & 195,665 & 10 & 666 \\
Synthetic Nonlinear & 6,000 & 3 & 3 \\
Synthetic Curved & 6,000 & 3 & 3 \\
\bottomrule
\end{tabular}
\caption{Summary of dataset statistics.}
\label{tab:datasets}
\end{table}

\subsection{Implementation Details}
\label{sec:implementation-details}
The data is split into $30\%$ for model fitting, $30\%$ for
policy training, $20\%$ for validation, and $20\%$ for testing.
Model fitting learns $f_{w_0}$, policy training runs Algorithm~\ref{therecoursealgorithm}, 
validation selects policy checkpoints
and baseline candidates (as we will describe shortly), and testing evaluates the
generalizability of selected policies on unseen samples. These are held-out splits, not cross-validation;
the experimental parameter grids are specified separately below.
We repeat this process 5 times using seeds $\{42,43,44,45,46\}$ with matched data and policy
We report average values and error bars (when applicable) in all experiments.

For the synthetic datasets, the ground-truth qualification scores are generated by our simulator 
and hence are known.
For real datasets, we learn the qualification scores from the binary-labeled dataset
and treat these learned scores as ground-truth. The learned proxy is trained on the
observed binary outcomes using binary cross-entropy: a $\tanh$ network
with two hidden layers (of size $(32,16)$) and a sigmoid activation function for the output.
We train the proxy using the Adam optimizer, early stopping on an internal $10\%$ holdout of the
model-fitting partition, and at most $1{,}500$ epochs. 

We learn two scoring rules $f_{w_0}$ from the generated ground-truth labels: one is an affine logit
and the other one is a quadratic logit.
Both scoring functions use soft-label binary cross-entropy and $\ell_2$ regularization
with a regularizer $\lambda=10^{-2}$, excluding the intercept.
They are fitted by L-BFGS-B. Affine coefficients and intercept are restricted to be 
in $[-5,5]$ and quadratic coefficients (for the quadratic logic) are restricted to be in $[10^{-4},3]$.
For real datasets, the mean AUC is approximately $0.87$ for ACSIncome, $0.91$-$0.92$
for Adult, $0.69$-$0.70$ for Give Me Some Credit, and $0.60$
for German Credit, with modest differences between scoring families.

We use Euclidean distance ($L^2$-norm) as the cost function $c$ when computing recourse. 
Immutable features cannot change during recourse generation. 
The policy uses separate affine sigmoid surrogates $\theta_1$ and $\theta_2$ for targets $q_{\theta_1}$ and
recommendations $\rho_{\theta_2}$, where $\theta_1,\theta_2\in\mathbb R^{d+1}$ and include an intercept. 
The non-intercept parameters are initialized independently from
$\mathcal N(0,0.02^2)$. The target and recommendation intercepts are
initialized to $-1$ and $0$, respectively. Both parameter vectors are
projected onto Euclidean balls of radius $3$.

Under the implementation's continuous relaxation of mutable coordinates,
computing the highest achievable score $\bar q$ in
Equation~\eqref{eq:q-bar} and the minimum-cost response
$\operatorname{br}$ in Equation~\eqref{eq:recourse-policy} is equivalent to a convex program 
when written using the score function's logit. We use closed-form solutions for
the affine logit and scalar multiplier bisection for the quadratic
logit. Derivatives of these responses with respect to the target are obtained
analytically or by implicit differentiation of the KKT equations,
allowing policy gradients to pass through the best response.

When running Algorithm~\ref{therecoursealgorithm}, we set the number 
projected full-gradient
updates to $T=500$. The competitive threshold $\bar{t}$ in Equation~\eqref{eq:smoothed-capacity} 
is recomputed using $80$ bisection steps at every update, and its implicit derivative
is included. The smooth recommendation weight mixes the outcomes of
no action and the complete response, rather than interpolating features.
A single $\tau$ smooths acceptance in both the objective and capacity
equation. 

Unless varied in an experiment, the default parameter settings are as follows:
\begin{table}[ht]
\centering
\begin{center}
\begin{tabular}{lll}
\toprule
Parameter & Default & Range of values tested \\
\midrule
$\lambda$ & $30$ & $14$ values in $[0.3,30]$ \\
$B$ & $0.75$ & $\{0.25,0.75,1.5\}$ \\
$\alpha$ & $0.4$ & $\{0.2,0.4,0.6\}$ \\
$\tau$ & $0.01$ & $\{0.005,0.01,0.02,0.05\}$ \\
$\eta$ & $0.2$ & $\{0.05,0.2\}$ \\
$T$ & $500$ & --- \\
\bottomrule
\end{tabular}
\end{center}
\caption{Default parameter setting.}
\label{tab:pamater-values}
\end{table}

In addition to Algorithm~\ref{therecoursealgorithm}, we use the following baselines:
\begin{enumerate}
\item \emph{No recourse:} where no individual changes features.
\item \emph{Original threshold:} where every eligible individual is provided with 
recourse to achieve the initial target of $t_0$. This tests the reliability
of initial-target recommendations under renewed competition, as
motivated by prior recourse-over-time studies~\citep{FonsecaBABS23}.
\item \emph{Common target:} where we perform a grid search to find a common target in $[t_0,1]$, 
for all individuals that achieves the lowest validation objective.
The recourse would be recommended to every rejected
individual who can attain that common target within the feasible budget of $B$.
\item \emph{Personalized targets without selection:} where we run Algorithm~\ref{therecoursealgorithm}
to learn $\theta_1$, while fixing $\rho=1$ for every eligible rejected
individual. 
\end{enumerate}

Cost and validity are averaged over all $N_-$ initially rejected individuals in $S^-$,
as in Equations~\eqref{eq:expected-recourse-cost} and
\eqref{eq:validity-definition}. Validity is their mean acceptance
probability, using expected acceptance under uniform cutoff tie-breaking.
Therefore, the upper bound on validity is $\min\{1,k/N_-\}$. However, this bound need not be
attainable under the recourse budget and policy restrictions. At the
population level, when the initially rejected fraction is $1-\alpha$,
the corresponding bound becomes $\min\{1,\alpha/(1-\alpha)\}$.

\begin{figure*}[ht!]
    \centering

    \begin{subfigure}[t]{0.85\textwidth}
        \centering
        \includegraphics[width=\linewidth]{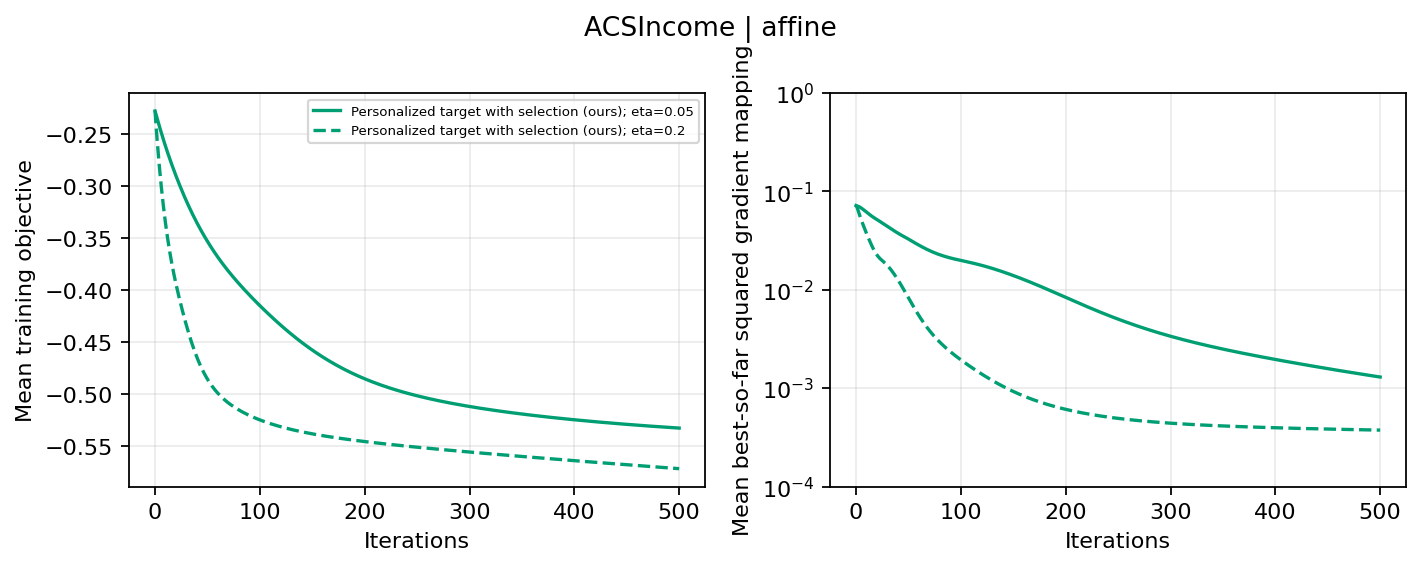}
        \caption{ASCIncome dataset, affine model, $\lambda=3$.\label{fig:convergence-asincome-affine-lambda-3}}        
    \end{subfigure}
    
    \hfill
    
    \begin{subfigure}[t]{0.85\textwidth}
        \centering
        \includegraphics[width=\linewidth]{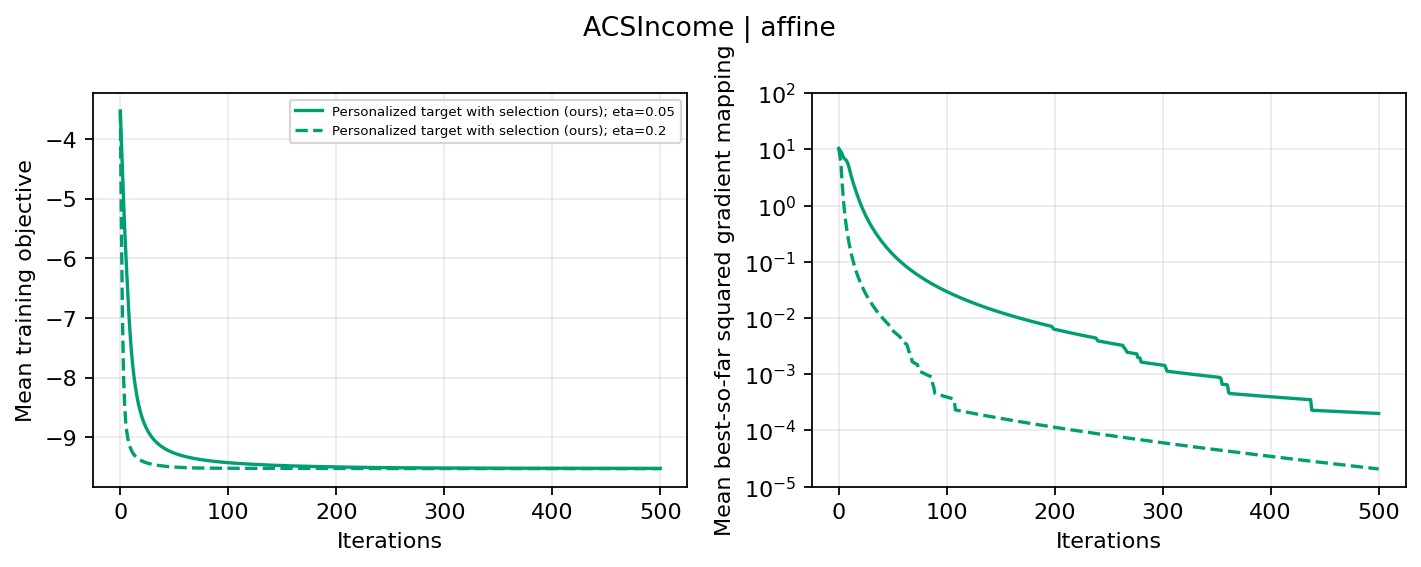}
        \caption{ASCIncome dataset, affine model, $\lambda=30$.\label{fig:convergence-asincome-affine-lambda-30}}
    \end{subfigure}
    
    \caption{Convergence results for the ASCIncome dataset with an affine scoring model. Each subfigure corresponds to a different $\lambda$ value as stated in the caption. Each curve corresponds to a different learning rate as indicated in the legend.\label{fig:convergence-asincome-affine}}
\end{figure*}

\subsection{Findings}
\label{sec:findings}
\paragraph{Convergence.}
In our first experiment, we study the convergence of
Algorithm~\ref{therecoursealgorithm}. Using the default parameter
settings, two values $\lambda\in\{3,30\}$, and two step sizes
$\eta\in\{0.05,0.2\}$, we run the algorithm for $T=500$ iterations.
We track the smooth empirical training objective and the best-so-far
squared projected-gradient mapping, averaged across
five seeds.

The results are presented in
Figure~\ref{fig:convergence-asincome-affine}
for the ACSIncome dataset with an affine scoring model as $f_{w_0}$.
Each subfigure corresponds to a different $\lambda$ value.
The left panel shows the training objective, evaluated on the
policy-training cohort, and the right panel shows the best-so-far
squared projected-gradient mapping, both as functions of the number
of iterations for both of the learning rates. Both quantities decrease substantially, with
$\eta=0.2$ producing faster progress than $\eta=0.05$ in these plots.
The objective stabilizes earlier for $\lambda=30$, although the
stationarity measure continues to improve. Because changing
$\lambda$ changes the objective itself, this observation does not
establish a general improvement in convergence rate.
Moreover, the best-so-far measure is non-increasing by construction, but
its reduction indicates that training finds iterates closer to
stationarity. The speed of improvement varies across datasets and scoring
functions. Additional results for other dataset and model pairs are provided in
Appendix~\ref{sec:app-exp-conv}.

\begin{figure}[ht!]
    \centering
    
    \begin{subfigure}[b]{0.48\textwidth}
        \centering
        \includegraphics[width=\textwidth]{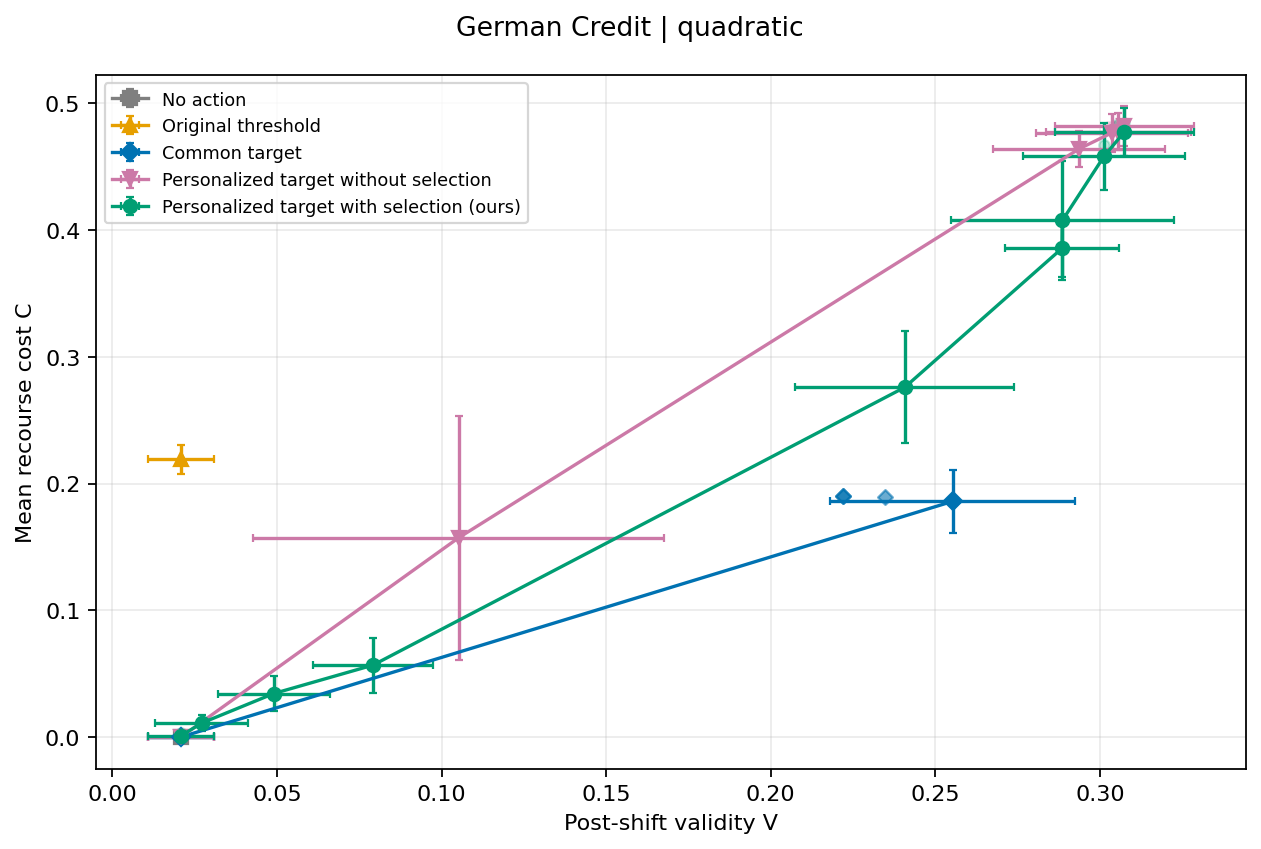}
        \caption{German Credit dataset, quadratic model.\label{fig:trade-off-german-credit-quadratic}}
    \end{subfigure}
    \hfill
    \begin{subfigure}[b]{0.48\textwidth}
        \centering
        \includegraphics[width=\textwidth]{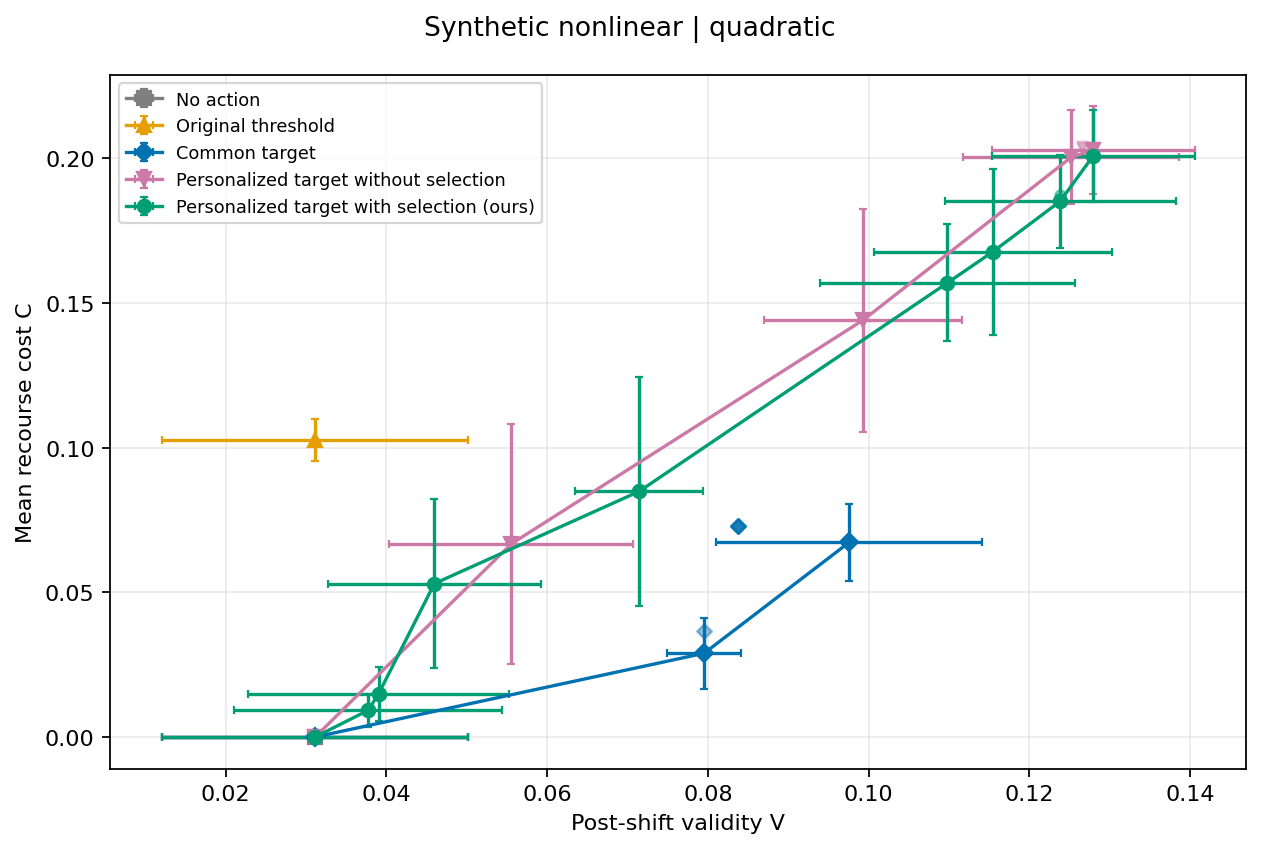}
        \caption{Synthetic Non-linear dataset, quadratic model.\label{fig:trade-off-synthetic-non-linear-quadratic}}
    \end{subfigure}
    
    \caption{The trade-off between recourse cost and post-shift validity. Each subfigure corresponds to a different dataset and scoring model as indicated by the caption. In each subfigure, each curve corresponds to the Pareto frontier of the trade-off between cost and validity for Algorithm~\ref{therecoursealgorithm} and the baselines. Dominated points remain visible as faint
markers.\label{fig:trade-off}}
\end{figure}

\paragraph{The Trade-Off Between Cost and Validity.}
We next study the trade-off between recourse cost and post-shift
validity for our algorithm and baselines. Across all datasets,
we vary $\lambda$ over 14 values between $0.3$ and $30$, while
keeping the other parameters at their default values.

The results are summarized in Figure~\ref{fig:trade-off}.
Each subfigure corresponds to a dataset and scoring model.
Cost and validity are computed over all initially rejected
applicants in each test cohort and then averaged across five seeds. For each method, the connected points show the nondominated cost-validity combinations for each of the evaluated policies.

We observe that no action incurs zero cost but can have nonzero measured validity. This is because the initial threshold $t_0$ is estimated from a reference population, whereas allocation is recomputed for each test cohort. But as expected, the validity achieved by no action is close to 0. 
Providing recourse with the original threshold incurs positive cost without significantly improving aggregate validity over the no-action baseline. This illustrates the limitation of recommending recourse to the
original threshold in competitive settings, an observation that has also been pointed out by prior work~\citep{FonsecaBABS23}.

Common targets offer favorable trade-offs at low to moderate
validity levels, and some observed personalized-policy points are 
dominated by common-target points. Personalized targets, with or without 
selection, nevertheless achieve the highest observed validity, albeit at higher cost. 
This is despite the observation that personalized targets achieve the lowest objective 
values in training (and often in testing), indicating greater validity does not
necessarily imply a lower objective value, perhaps due to overfitting. 

Follow-up target diagnostics help explain these differences.
At $\lambda=3$, personalized policies with selection assign higher targets 
than the common-target policy to approximately
$80\%$-$100\%$ of applicants receiving recommendations under both methods, 
depending on the dataset and scoring function. Their mean recourse
cost ranges from $0.695$ to $0.745$, close to the budget $B=0.75$,
compared with $0.507$-$0.576$ for common targets.
Selection reduces expenditure on unsuccessful actions relative
to implementing the same learned targets for every eligible
applicant, indicating that the learned target levels and recipient 
selection both contribute to the observed trade-offs. 

Analogous to the results in the robust recourse literature~\cite{UpadhyayJL21, NguyenBN+22, KyawKJ26, PawelczykDHKL23},
our results show that achieving high validity under competition also imposes a significant increase in 
the cost of implementing the recourse.
Additional results for other dataset and model pairs are provided in 
Appendix~\ref{sec:app-exp-trade-off}, where many of our observations continue to hold.

\paragraph{Effect of Parameters.}
To better understand the effect of each of the parameters in our setting, we 
conducted sensitivity analysis on the choice of parameters through three different
experiments: (1) In budget sensitivity experiments, we varied budget from 0.25 to 1.75
for 3 choices of $\lambda \in \{3, 10, 30\}$ while keeping the other parameters 
at their default values. (2) In admission rate sensitivity experiments, we varied $\alpha$ 
from 0.2 to 0.6 for the same three values of $\lambda$ and other default parameters. 
(3) In temperature sensitivity experiments, we selected $\tau$ values ranging from 0.005 to
0.005 for the same three values of $\lambda$ and other default parameters. 

\begin{figure}[ht!]
    \centering
    \includegraphics[width=\linewidth]{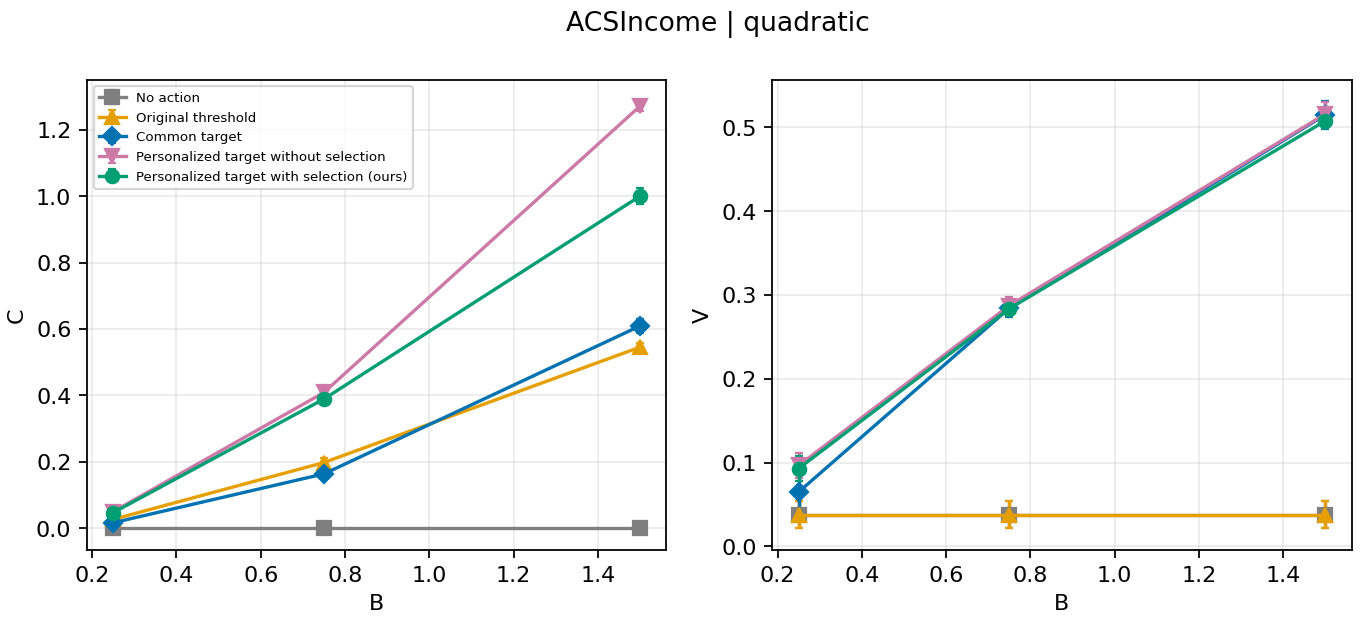}
    \caption{The effect of budget $B$ on cost and validity of recourse in the ACSIncome dataset for quadratic cost at $\lambda=10$. Curves correspond to different approaches.\label{fig:budget-sensitivity}}
\end{figure}

\begin{figure}[ht!]
    \centering
    \includegraphics[width=\linewidth]{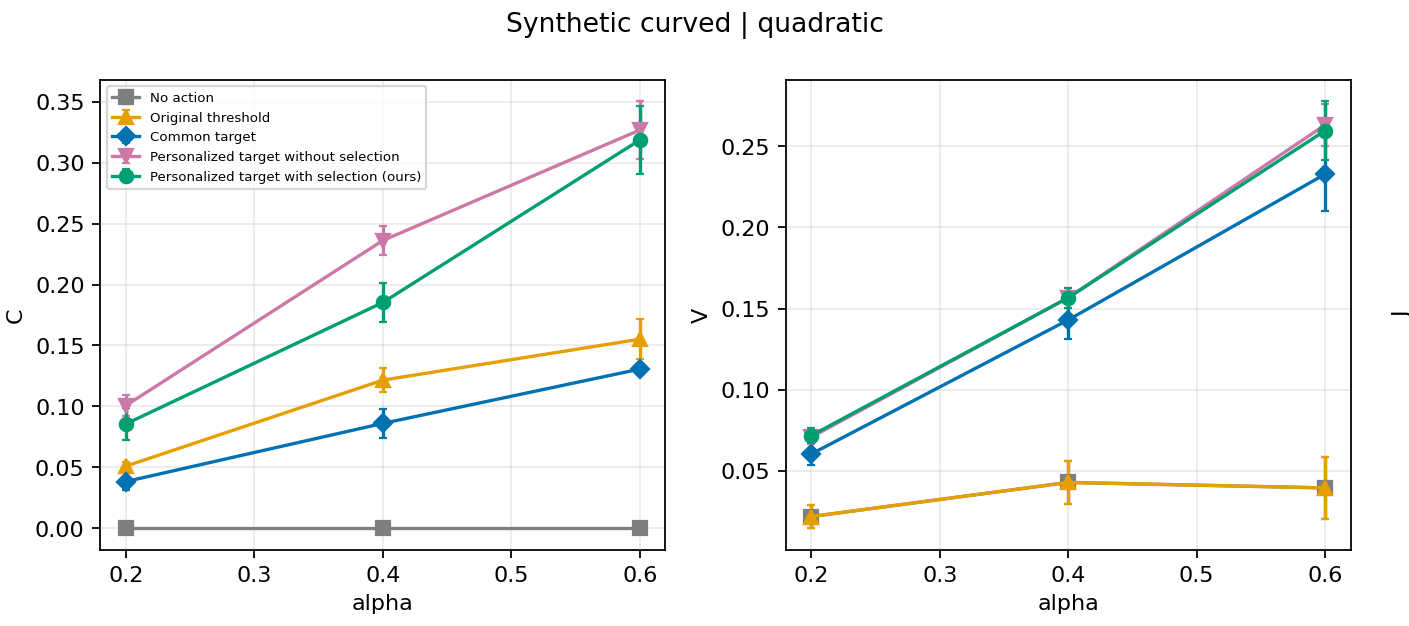}
    \caption{The effect of admission rate $\alpha$ on cost and validity of recourse in the Synthetic Curved dataset for quadratic cost at $\lambda=30$. Curves correspond to different approaches.\label{fig:alpha-sensitivity}}
\end{figure}

\begin{figure}[ht!]
    \centering
    \includegraphics[width=0.5\linewidth]{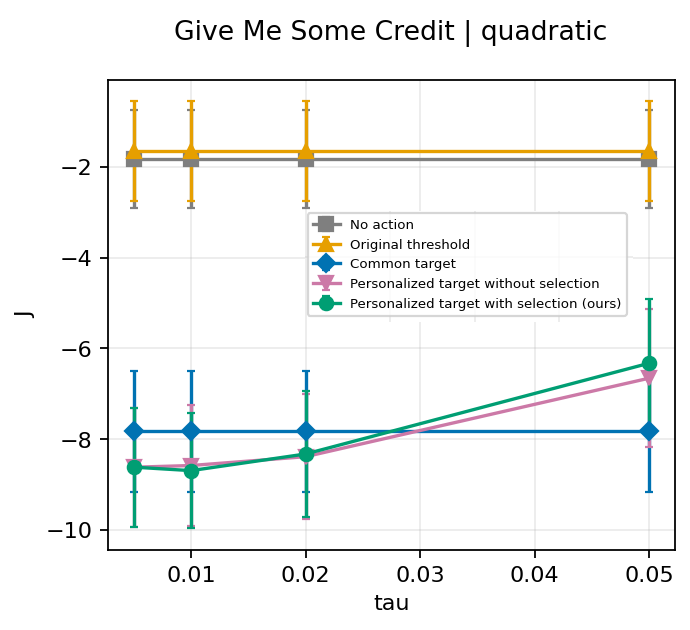}
    \caption{The effect of temperature $\tau$ on the recourse objective in the Synthetic Curved dataset for quadratic cost at $\lambda=30$. Curves correspond to different approaches.\label{fig:temp-sensitivity}}
\end{figure}

Increasing the budget generally improves post-shift validity,
but also increases the cost. For example, on ACSIncome with a quadratic
scoring function and $\lambda=10$, as depicted in Figure~\ref{fig:budget-sensitivity}, 
increasing the budget $B$ from $0.25$ to $1.5$
raises Algorithm~\ref{therecoursealgorithm}'s mean validity from $9.6\%$ to $51.4\%$
and its mean cost from $0.05$ to $1.11$.
However, additional expenditure does not necessarily provide much advantage: 
at the budget $B=1.5$, the common-target baseline
achieves approximately the same validity at a mean cost of $0.61$.
Larger budgets, therefore, expand the opportunities for successful
recourse, but, by themselves, do not ensure efficient use of those
opportunities.

Increasing the admission rate consistently raises Algorithm~\ref{therecoursealgorithm}'s mean validity
across all tested datasets, models, and $\lambda$ combinations.
For Synthetic Curved with a quadratic scoring function and $\lambda=30$, 
as depicted in Figure~\ref{fig:alpha-sensitivity},
increasing $\alpha$ from $0.2$ to $0.6$ raises validity from
$7.1\%$ to $26.3\%$, while increasing the mean cost from $0.10$ to $0.32$.
In general, greater capacity increases the average recourse expenditure as
the eligible population, and selected
recommendations will also change with the budget increase.

Performance is relatively stable across the tested temperatures
for many dataset and model pairs, but excessive smoothing sometimes can
substantially worsen the achieved recourse objective in Equation~\eqref{eq:validity}.
On Give Me Some Credit with a quadratic scoring function, as depicted in Figure~\ref{fig:temp-sensitivity}, 
increasing the temperature $\tau$ from $0.01$ to $0.05$ worsens Algorithm~\ref{therecoursealgorithm}'s objective
value from $-8.62$ to $-6.28$ by reducing both mean validity and cost.
However, the results are nearly identical for smaller $\tau$ values.
These findings support our approach of selecting the temperature through
validation, as the effect depends on the dataset and scoring function.

\paragraph{Target Distributions.}
We next examine how common and personalized targets affect the
score improvements and costs required of recourse recipients.
We use $\lambda\in\{1,3,10\}$ and default values for all the other parameters.

\begin{figure*}[ht!]
    \centering
    \includegraphics[width=\linewidth]{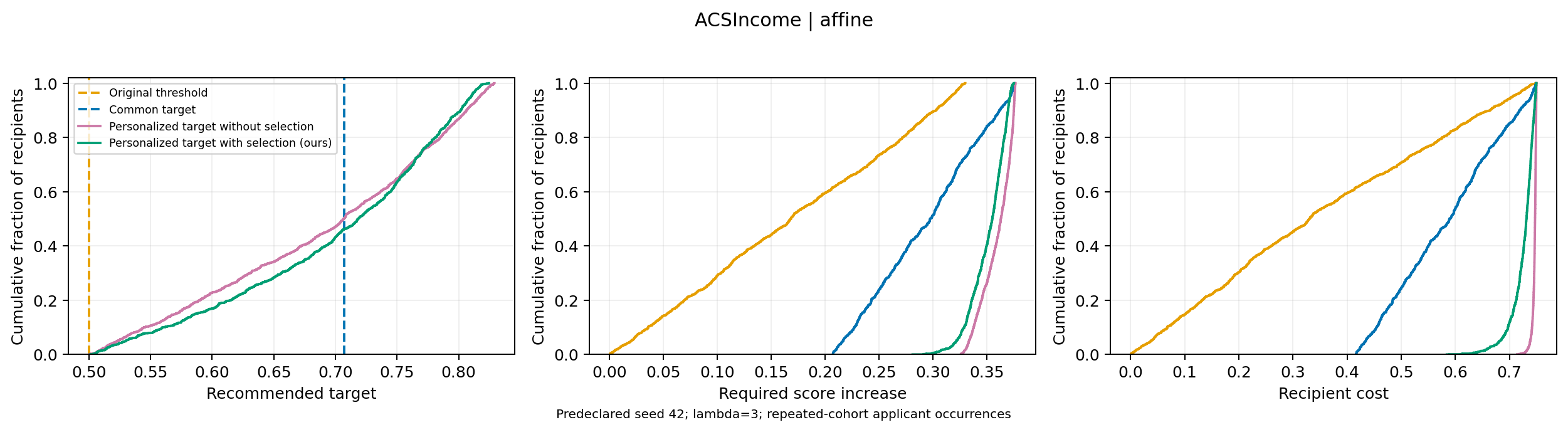}
    \caption{Target distributions for the ACSIncome dataset with an affine scoring
    model and $\lambda=3$. The panels show the cumulative density function of
    the recommended targets (left),
    required score increases (middle), and recourse costs (right)
    among recipients for seed $42$.
    Vertical dashed lines in the left panel indicate the original
    threshold and common target.
    \label{fig:disp-2}}
\end{figure*}

Figure~\ref{fig:disp-2} shows results for the ACSIncome dataset with an affine
scoring model and $\lambda=3$. The left panel shows recommended
scores $q(x)$, the middle panel shows required score increases
$q(x)-f_{w_0}(x)$, and the right panel shows the minimum cost of
attaining the recommended target. The curves give the cumulative
fraction of recipient observations whose value is at most the
horizontal-axis value. The middle and right panels distinguish
improvement in score from effort measured by the recourse cost to achieve the 
improved score.

The left panel of Figure~\ref{fig:disp-2} shows that both personalized
policies assign very similar targets, and the fraction of individuals receiving 
each target score is almost uniform. 
The middle panel shows that even a common target requires
different score increases because recipients have different
initial scores. Furthermore, personalized policies generally
require larger score increases than the common targets.
The right panel shows that personalization does not necessarily
reduce effort. The median recipient cost is approximately $0.589$
for the common target, $0.747$ for personalized targets without
selection, and $0.731$ for personalized targets with selection.
Thus, many personalized recommendations require expenditure close
to the maximum allowable budget of $B=0.75$.
Further analysis, comparing the personalized policy with
common-target policy, indicates that the personalized policy with
selection assigns higher targets than the common-target policy
to approximately $94\%$ of their shared recipient,
with a mean additional cost of $0.147$.

\begin{figure*}[ht!]
    \centering
    \includegraphics[width=\linewidth]{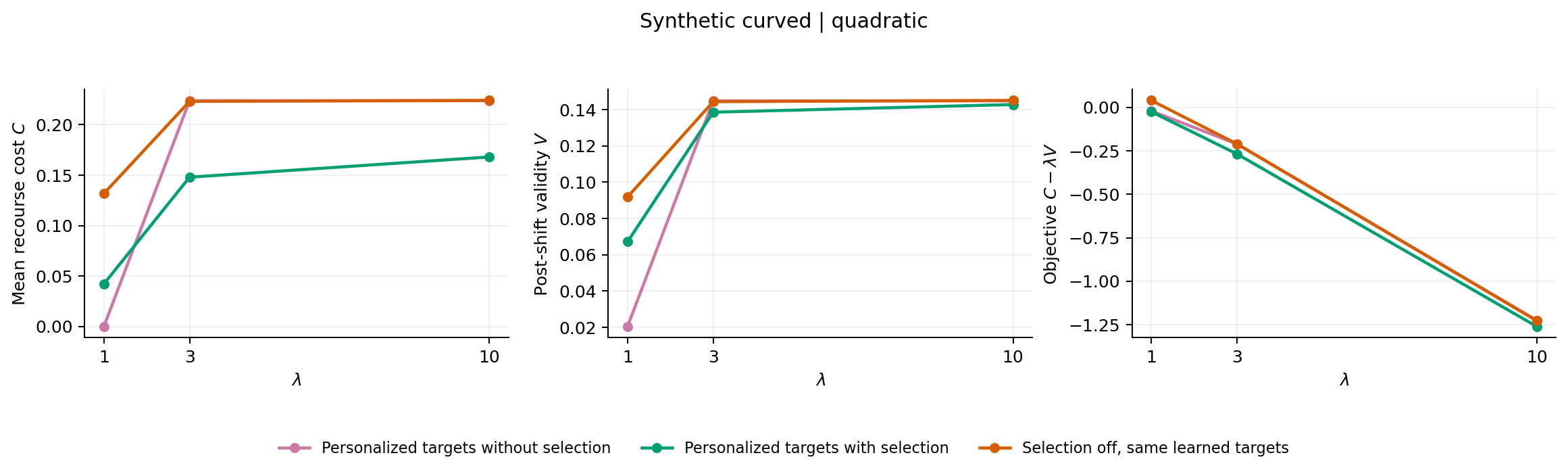}
    \caption{Effect of modifying the recipient selection for the Synthetic Curved dataset
    with a quadratic scoring model. The panels report mean cost
    (left), post-shift validity (middle), and
    recourse objective (right) for $\lambda\in\{1,3,10\}$.
    \label{fig:disp-3}}
\end{figure*}

\paragraph{The Role of Selection.}
To isolate the effect of recipient selection, we take a learned
personalized policy, retain its target function $q$, and replace
its recommendation rule with $\rho(x)=1$ for every eligible rejected
applicant. Previously selected applicants retain their recommended
actions, while previously unselected eligible applicants now
implement their minimum-cost responses to the learned targets.
We recompute the competitive threshold and measure cost,
validity, and the objective value. In this experiment, we use $\lambda\in\{1,3,10\}$ 
and default values for all the other parameters.

\begin{figure*}[ht!]
    \centering
    \includegraphics[width=\linewidth]{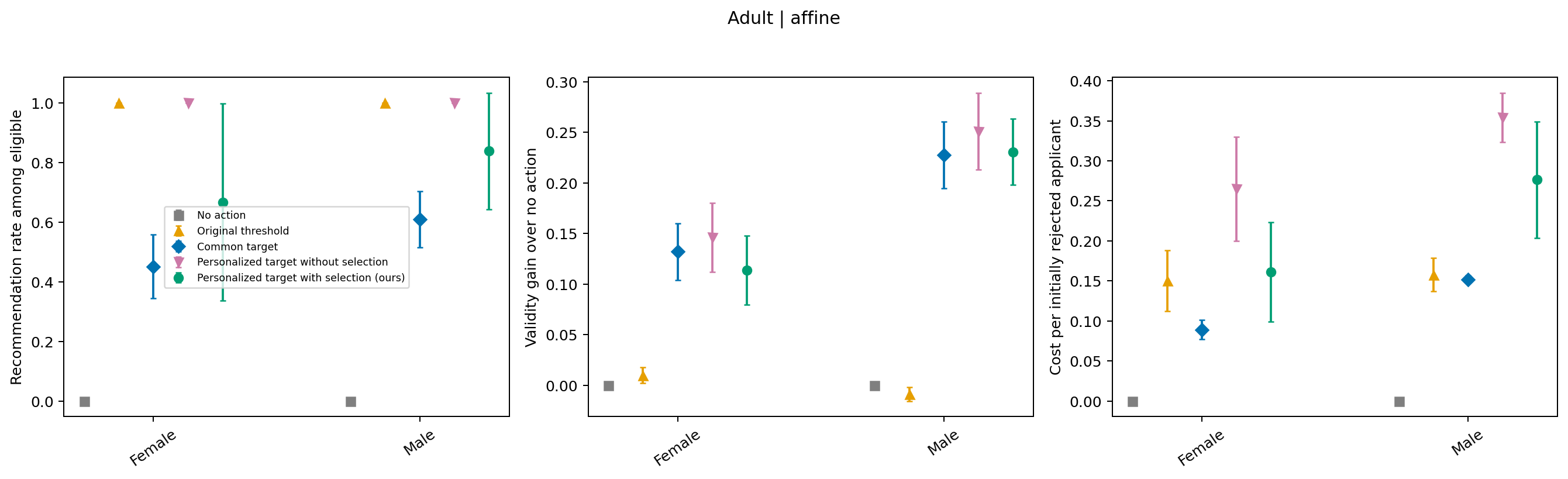}
    \caption{Group outcomes partitioned by gender for the Adult dataset with an affine
    scoring model and $\lambda=3$. The panels report recommendation
    rates among eligible rejected applicants (left), post-shift
    validity gains relative to no action (middle), and mean cost
    per initially rejected applicant (right).
    Points show means across the seeds with
    $95\%$ confidence intervals.
    \label{fig:disp-1}}
\end{figure*}

Figure~\ref{fig:disp-3} illustrates this comparison for the Synthetic
Curved dataset with a quadratic scoring function.
The three panels plot outcomes against $\lambda$.
The left panel reports mean recourse cost (including expenditure
on unsuccessful recourse). The middle panel reports post-shift
validity. The right panel reports the objective value as in Equation~\eqref{eq:validity}.
At $\lambda=3$, disabling selection increases mean cost from
approximately $0.148$ to $0.223$ (left panel), while validity
increases only from $0.139$ to $0.144$ (middle panel).
Consequently, the objective worsens from approximately $-0.268$
to $-0.210$ (right panel).
Further analysis shows that expenditure
on unsuccessful recourse increases from $0.050$ to $0.121$ per
initially rejected applicant. Thus, selection
can avoid substantial expenditure for a small reduction in
aggregate validity.

\begin{figure*}[t]
    \centering
    \includegraphics[width=\linewidth]{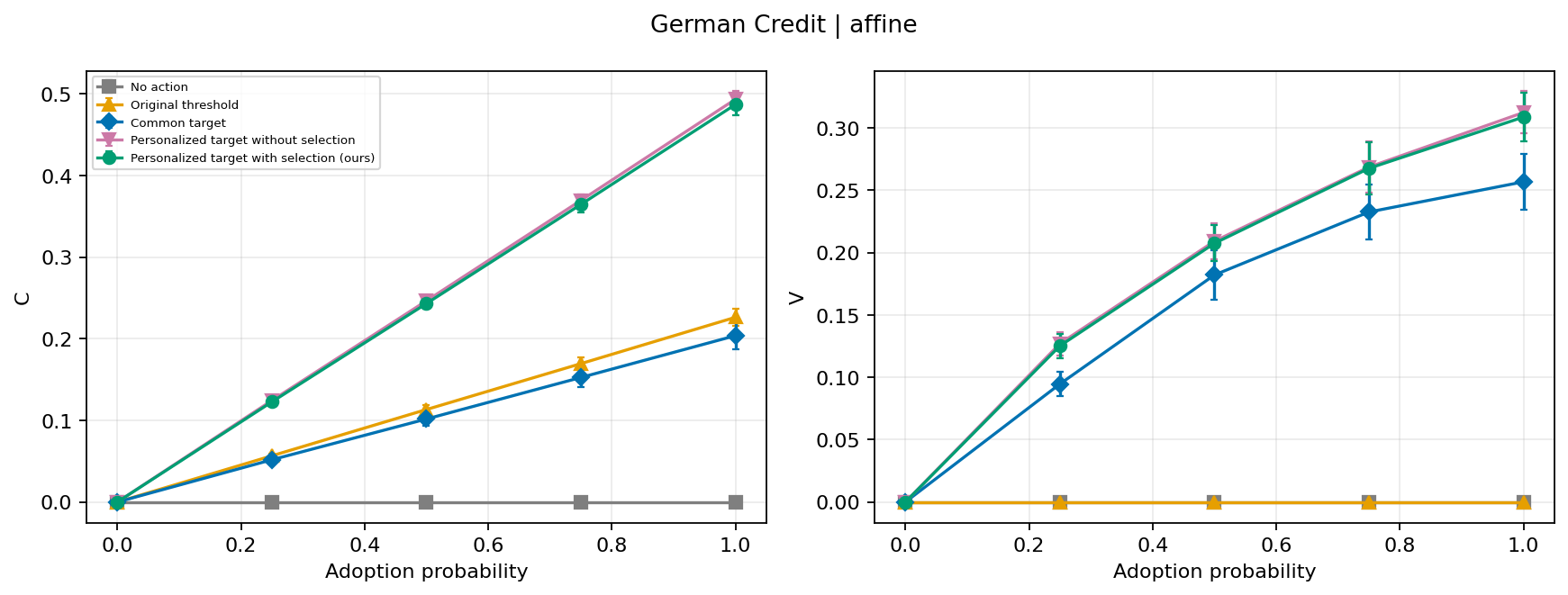}
    \caption{Partial adoption for the German Credit dataset with an affine
    scoring model and $\lambda=30$. The policy selected under full
    adoption is held fixed while the adoption probability varies,
    and the competitive allocation is recomputed after each
    realization.}
    \label{fig:partial-adoption}
\end{figure*}

\paragraph{Group Outcomes.}
We also examine how recourse benefits and costs are distributed
across subpopulations. In this experiment, we use $\lambda\in\{1,3,10\}$ 
and default values for all the other parameters. To form the subpopulations, 
we used gender or race for real datasets. For synthetic datasets,
we use diagnostic groups defined by the sign of the immutable
first coordinate.

Figure~\ref{fig:disp-1} shows outcomes by gender for the Adult dataset
with an affine scoring model and $\lambda=3$.
The left panel reports recommendation rates among eligible rejected
applicants. The middle panel reports validity gains relative to
no action. The right panel reports mean cost over initially
rejected members of each group.
For personalized targets with selection, the mean recommendation
rate is approximately $83.9\%$ among eligible male applicants
and $66.7\%$ among eligible female applicants.
Validity gains are approximately $23.1\%$ and $11.4\%$, 
respectively, while mean costs are $0.276$ and $0.161$.

These results reveal differences in both access to recommendations
and realized benefits. The lower mean cost for female applicants
does not imply that their recommended actions are individually
cheaper: cost per initially rejected applicant also reflects how
many applicants receive recommendations.
The comparisons show that optimizing aggregate
cost and validity does not ensure equal group outcomes.  This is not 
surprising as equity is not included in the optimization objective~\citep{UstunSL19}.

\paragraph{Partial Adoption.}
We examine how incomplete adoption affects policies designed under
the assumption that every recommendation is implemented. We keep the
policy selected under full adoption fixed and let each recommended
applicant independently implement their response with probability
$p$ where $p$ belongs to the set $\{0,0.25,0.5,0.75,1\}$.
Non-adopters retain their original
features, and we recompute the competitive allocation after each
realization. We measure cost and validity over all
initially rejected applicants. 

Figure~\ref{fig:partial-adoption} presents results for the German Credit dataset
with an affine scoring model, $\lambda=30$, and the remaining
parameters at their default values. At $50\%$ adoption, mean cost
is approximately $0.246$ and validity is $20.9\%$, compared with
$0.493$ and $31.2\%$ under full adoption. Thus, half adoption retains
approximately two-thirds of full-adoption validity at half the cost.
Although expected expenditure scales proportionally with adoption
when recommendations remain fixed, validity need not. This is because fewer
implemented actions also change the competition faced by adopters.

Further analysis across all $12$ datasets and scoring function
combinations shows that, at $\lambda=30$, average validity increases
as more applicants implement their recommendations. Full adoption
achieves the lowest average objective value: although more
applicants incur recourse costs, the increase in validity outweighs
this additional expenditure at the chosen value of $\lambda$.
Therefore, partial adoption can preserve some of the benefits of
recourse, while reducing both expenditure and overall validity.
\section{Discussion and Limitations}
\label{sec:disc}
In this work, we introduced a framework for algorithmic recourse under
competition that explicitly anticipates the effects of population-wide
feature modifications. We studied population-wide recourse computation policies
and how these policies can affect the validity and cost of recourse.

We discuss some of the limitations of our setting and flesh out new directions
for future work.
First, our analysis assumes that achieved scores and recourse costs
depend smoothly on the policy parameters. Even in such settings, our approach
does not guarantee optimality. Understanding under what conditions optimal strategies 
can be designed for the smooth setting is a natural next step. Handling nonsmooth response
changes and analyzing the performance in such settings likely requires different optimization 
techniques and analysis. 
Second, our formulation
models a single round of recourse in which (in almost all our experiments) individuals fully implement the recourse. 
Extending the framework to a repeated setting where individuals repeatedly go through the 
decision-making process, adoption of recourse is selective and asynchronous
remains an important direction. Finally, improvements in aggregate cost and validity do 
not mean these benefits are distributed equitably across all individuals and subpopulations. 
Understanding what it means for a recourse policy to be equitable in a competitive setting
and providing equitable recourse policies for such settings is left for future work
(see the Ethical Consideration section).

\paragraph{Ethical Consideration.}
Our framework assumes that the decision-maker's goal is to help initially
rejected individuals by balancing their aggregate recourse cost and
post-shift validity. Although we model individual costs and competitive
outcomes, we do not assess the broader consequences
of these decisions for individuals and subpopulations. Selecting recipients
may distribute acceptance opportunities unequally across subpopulations, unsuccessful recourse may impose
unrewarded effort, and initially accepted individuals may lose access
to the resource due to future competition. Since our objective does not incorporate 
equity into considerations, improvements in aggregate cost and validity should not
be interpreted as guarantees of individual/subpopulation benefit or societal welfare.
Future work should carefully analyze these aspects.

\paragraph{Acknowledgments.} We thank Kshitij Kayastha for discussions during the early stages of this work. 

\paragraph{Usage of LLMs.}
Gemini 3.1 Pro was used to write the initial version of the experiments from the problem formulation
and algorithm. GPT-6 Astra was used to verify the correctness of implementation. We then checked the edited code by Astra. 
GPT-6 Astra was used to proofread the paper and provide edits, especially in the implementation details of the 
experiments. LLMs were not used for problem formulation, formation of research questions, algorithmic solutions, and 
the description of related work.

\bibliographystyle{plainnat}
\bibliography{bib}

\appendix
\section{Additional Experimental Results}
\label{sec:app-exp}

\subsection{Convergence}
\label{sec:app-exp-conv}

\begin{figure}[ht!]
    \centering

    \begin{subfigure}[t]{0.49\textwidth}
        \centering
        \includegraphics[width=\textwidth]{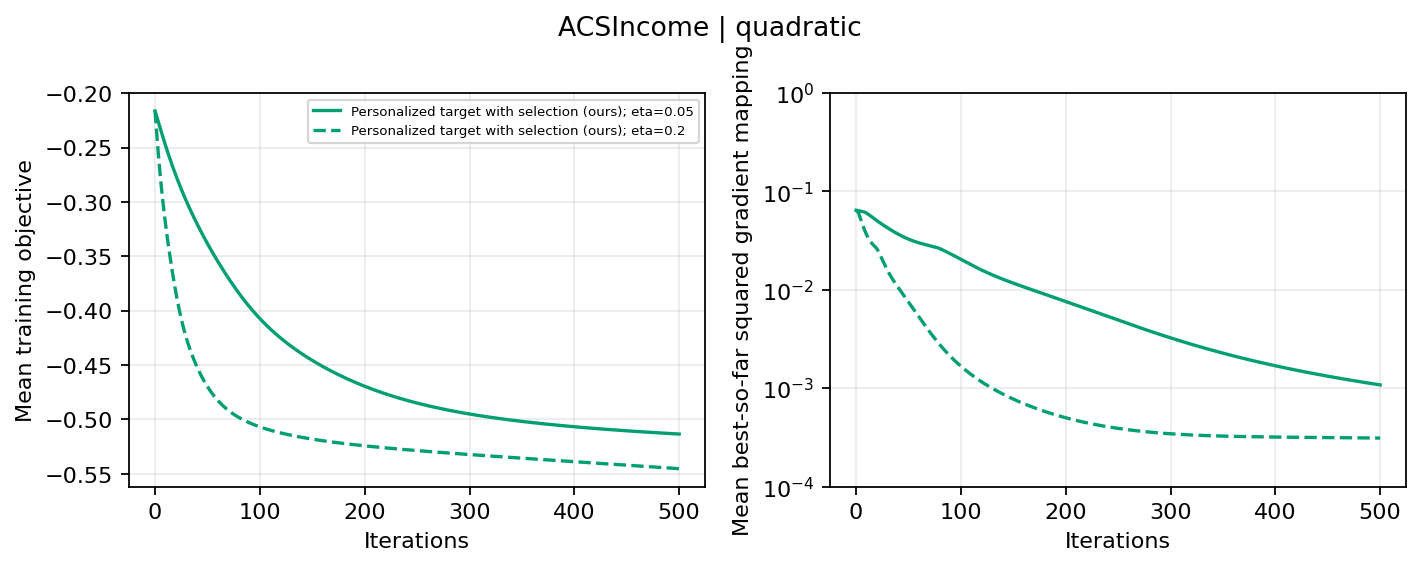}
        \caption{ASCIncome dataset, quadratic model, $\lambda=3$.\label{fig:convergence-asincome-quadratic-lambda-3}}        
    \end{subfigure}
    \hfill
    \begin{subfigure}[t]{0.49\textwidth}
        \centering
        \includegraphics[width=\textwidth]{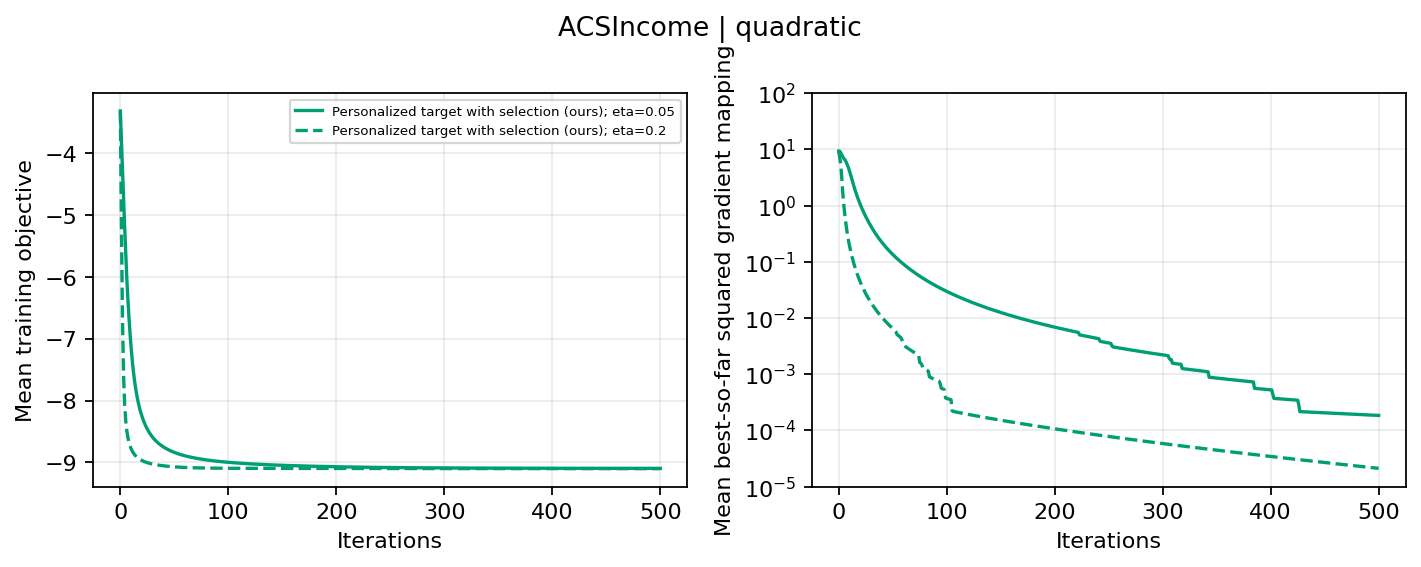}
        \caption{ASCIncome dataset, quadratic model, $\lambda=30$.\label{fig:convergence-asincome-quadratic-lambda-30}}
    \end{subfigure}
    
    \caption{Convergence results for the ASCIncome dataset with a quadratic scoring model. Each subfigure corresponds to a different $\lambda$ value as stated in the caption. Each curve corresponds to a different learning rate as indicated in the legend.\label{fig:convergence-asincome-quadratic}}
\end{figure}

\begin{figure}[ht!]
    \centering

    \begin{subfigure}[t]{0.49\textwidth}
        \centering
        \includegraphics[width=\textwidth]{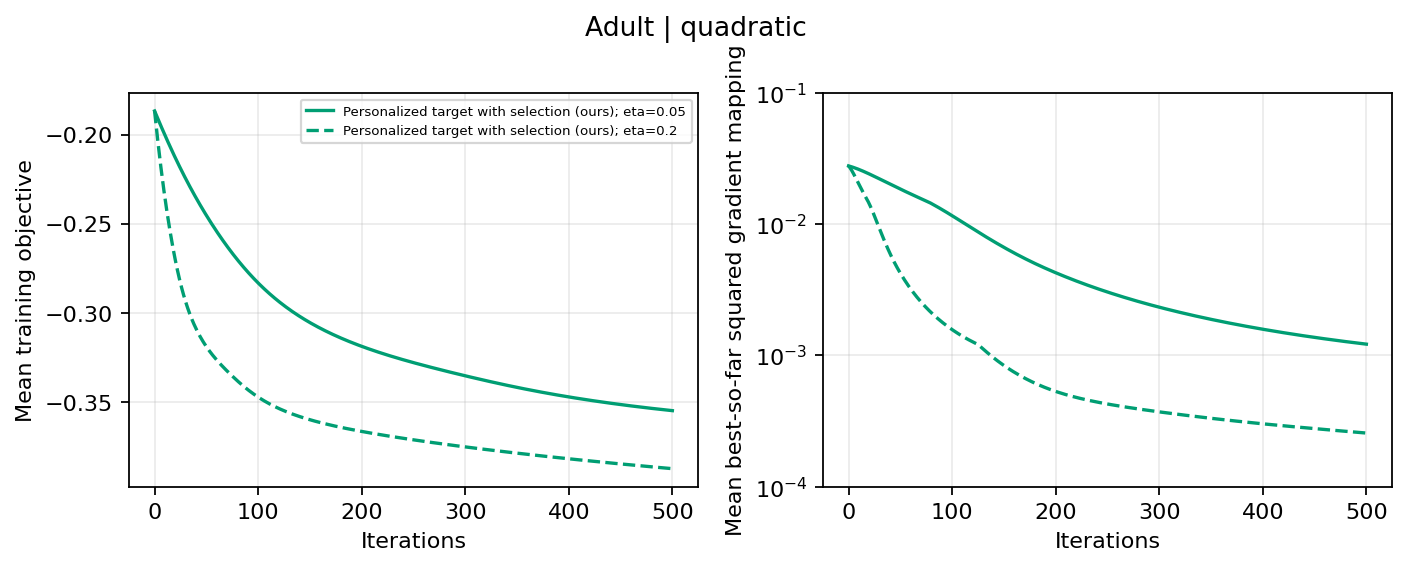}
        \caption{ASCIncome dataset, quadratic model, $\lambda=3$.\label{fig:convergence-adult-quadratic-lambda-3}}
    \end{subfigure}
    \hfill
    \begin{subfigure}[t]{0.49\textwidth}
        \centering
        \includegraphics[width=\textwidth]{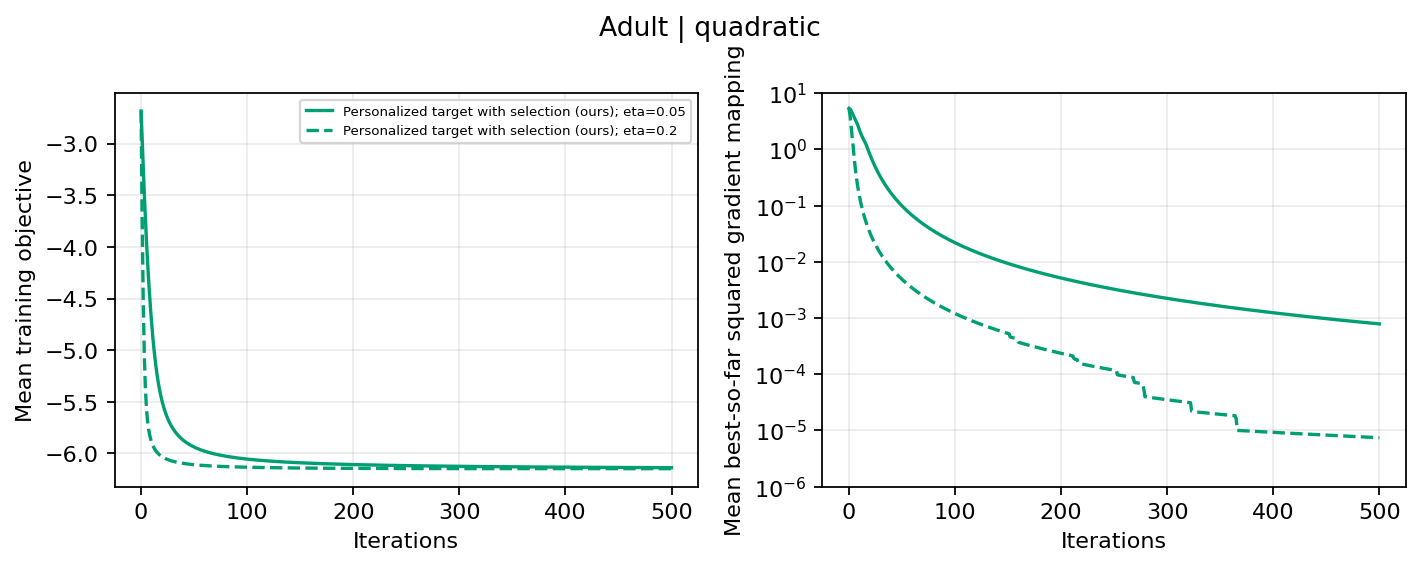}
        \caption{ASCIncome dataset, quadratic model, $\lambda=30$.\label{fig:convergence-adult-quadratic-lambda-30}}
    \end{subfigure}
    
    \caption{Convergence results for the Adult dataset with a quadratic scoring model. Each subfigure corresponds to a different $\lambda$ value as stated in the caption. Each curve corresponds to a different learning rate as indicated in the legend.\label{fig:convergence-app}}
\end{figure}

\subsection{The Trade-Off Between Cost and Validity}
\label{sec:app-exp-trade-off}
\begin{figure}[ht!]
    \centering
    
    \begin{subfigure}[b]{0.45\textwidth}
        \centering
        \includegraphics[width=\textwidth]{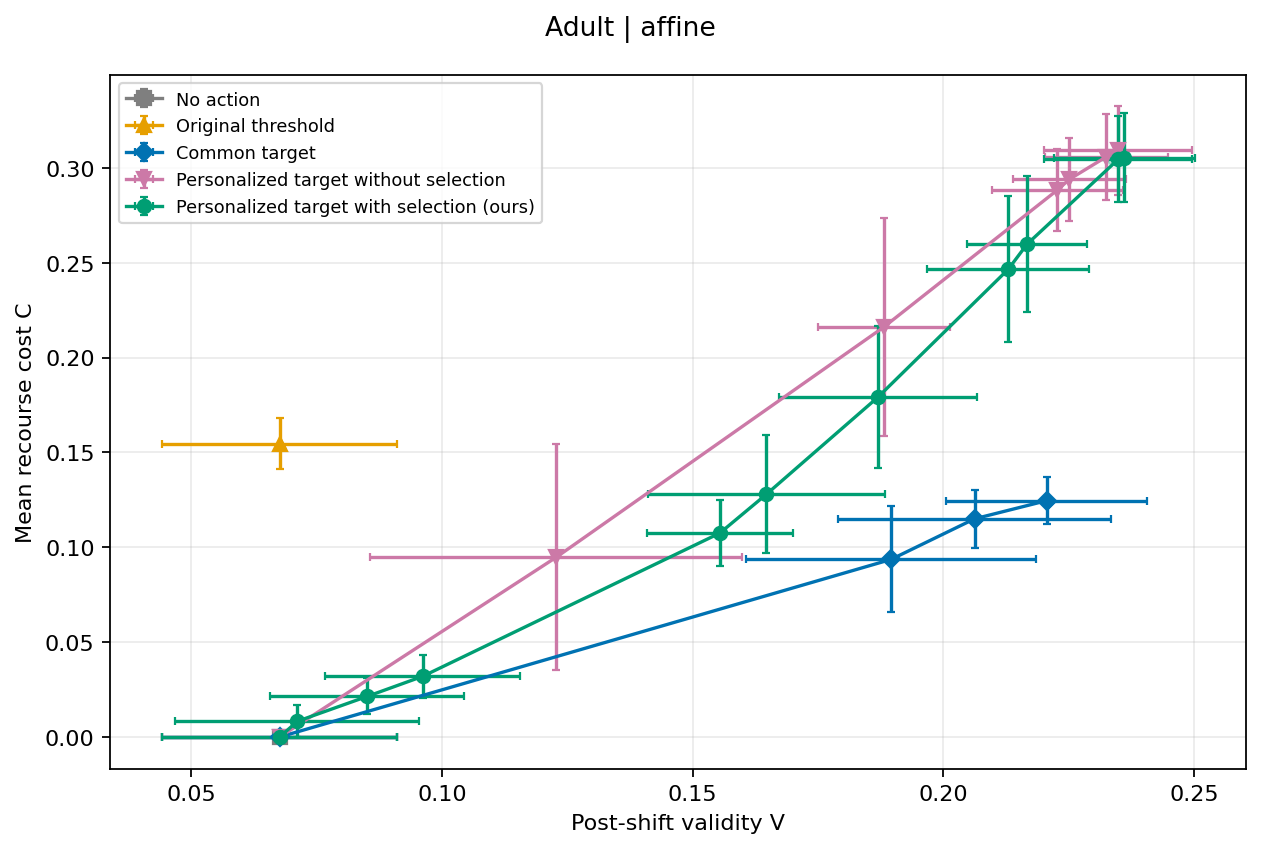}
        \caption{Adult dataset, affine model.\label{fig:trade-off-adult-affine}}
    \end{subfigure}
    \hfill
    \begin{subfigure}[b]{0.45\textwidth}
        \centering
        \includegraphics[width=\textwidth]{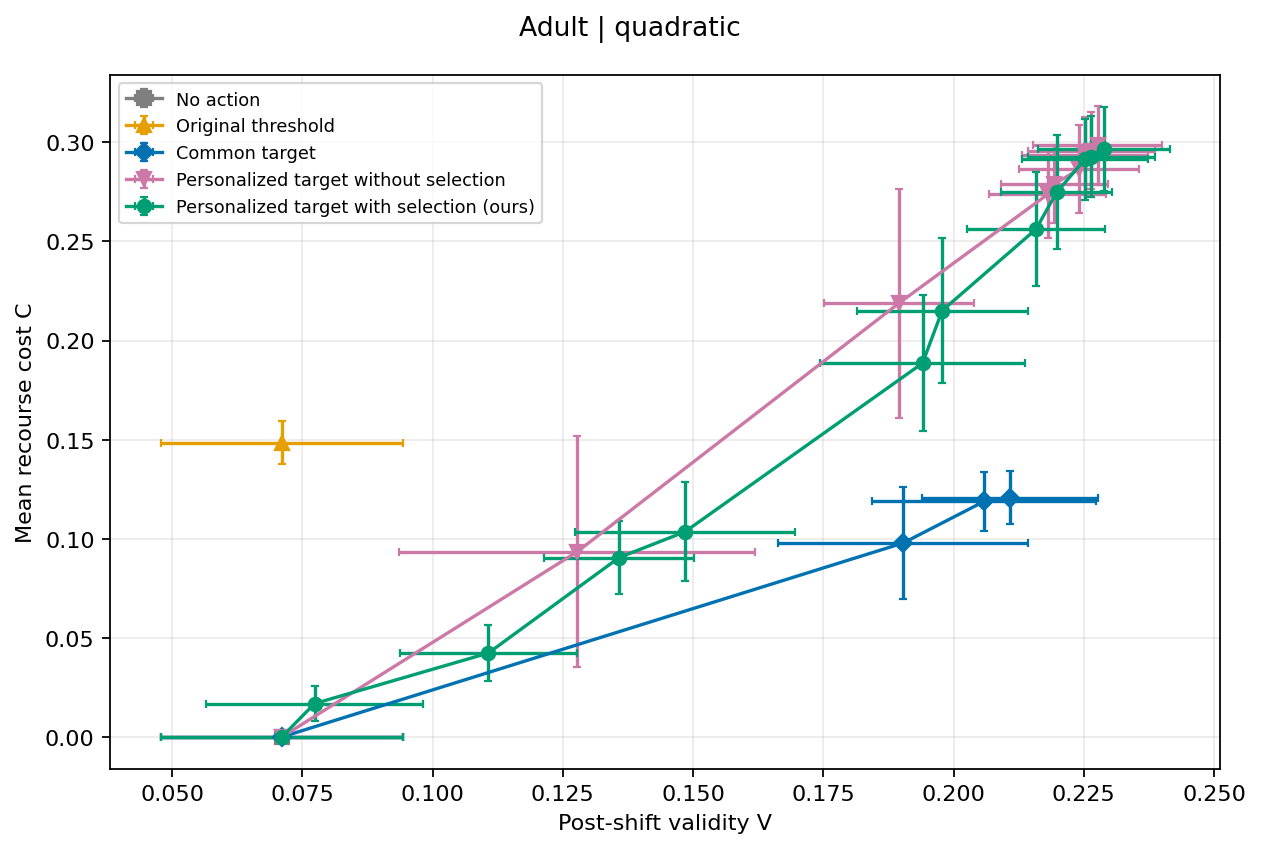}
        \caption{Adult dataset, quadratic model.\label{fig:trade-off-adult-quadratic}}
    \end{subfigure}
    \hfill
    \begin{subfigure}[b]{0.45\textwidth}
        \centering
        \includegraphics[width=\textwidth]{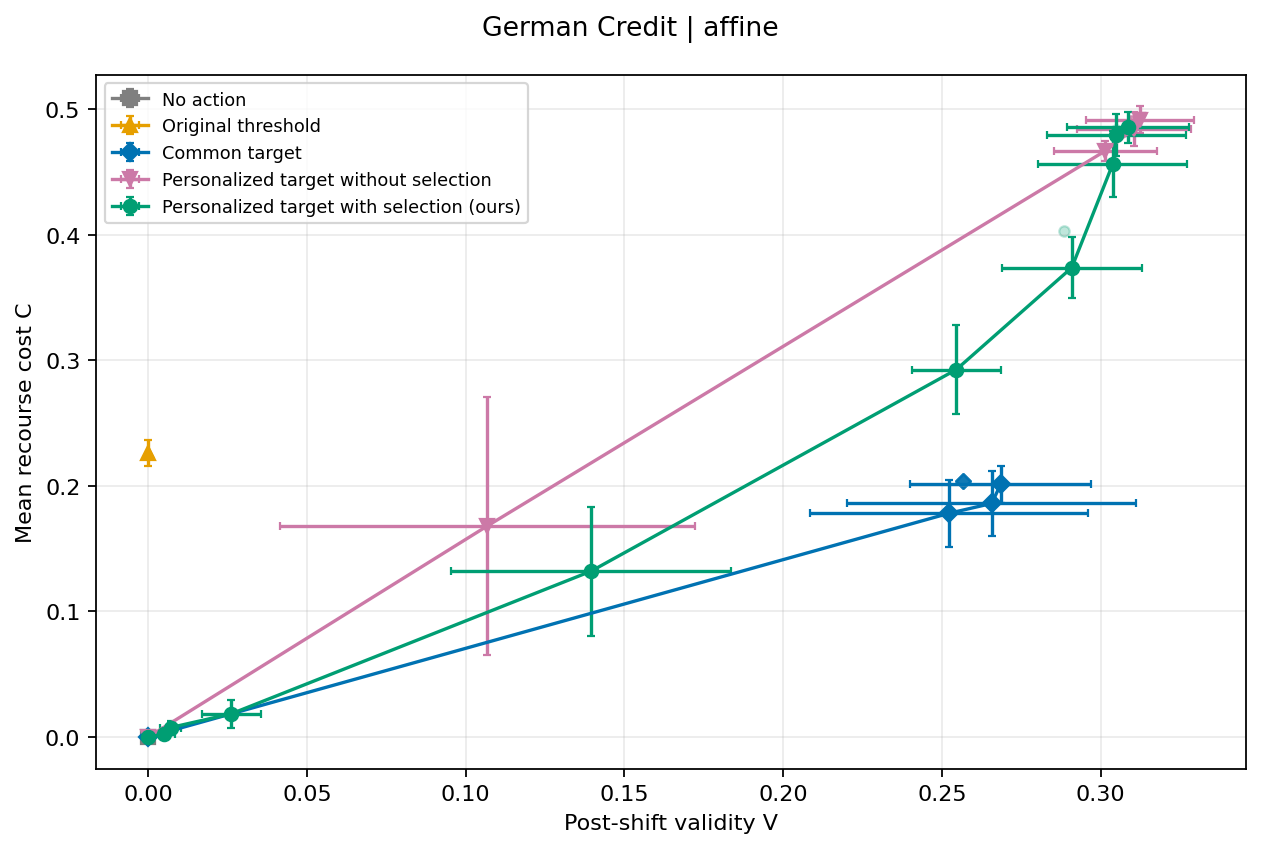}
        \caption{German Credit dataset, affine model.\label{fig:trade-off-german-quadratic}}
    \end{subfigure}
    \hfill
    \begin{subfigure}[b]{0.45\textwidth}
        \centering
        \includegraphics[width=\textwidth]{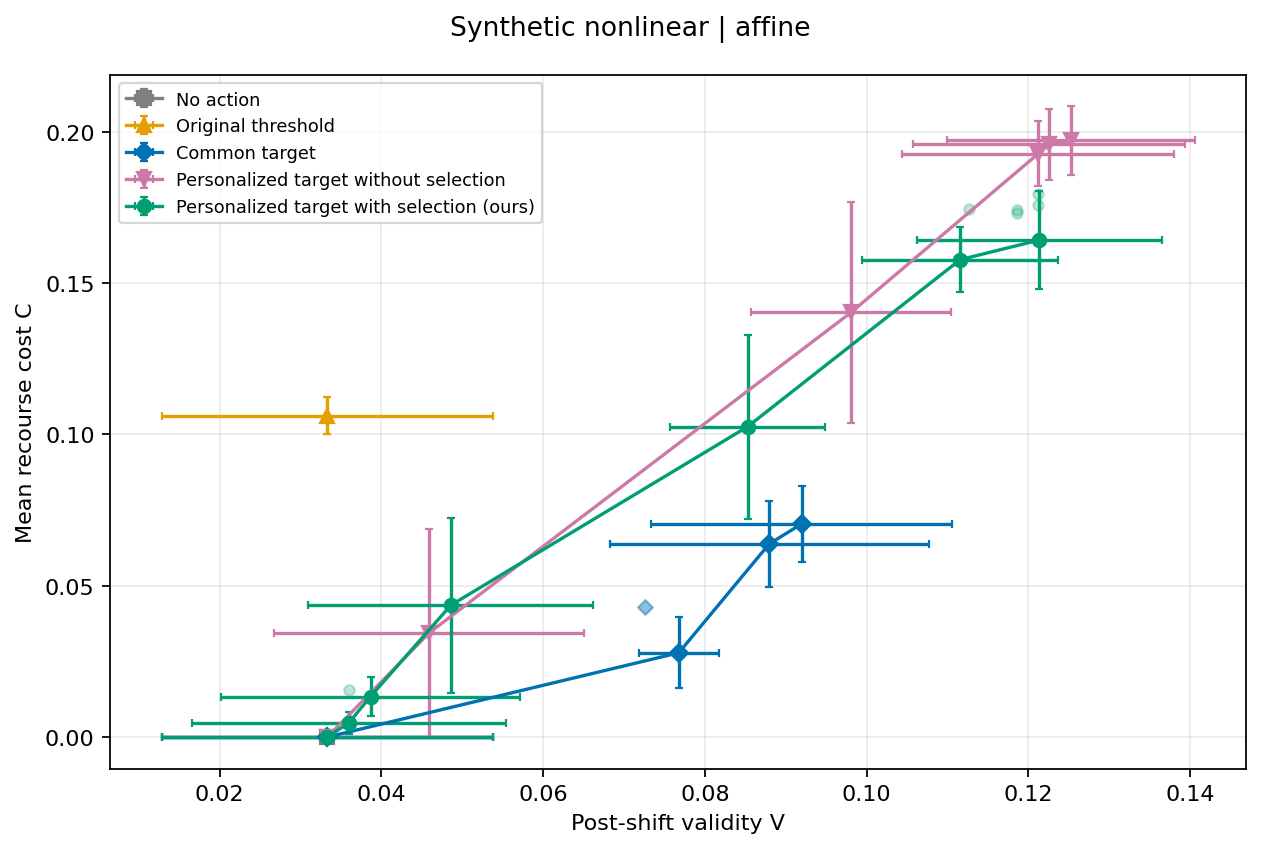}
        \caption{Synthetic non-linear dataset, affine model.\label{fig:trade-off-synsthetic-non-linear-affine}}
    \end{subfigure}

    \caption{The trade-off between validity of recourse and its cost. Each subfigure corresponds to a different dataset and scoring model as indicated by the caption. In each subfigure, each curve corresponds to the Pareto frontier of the trade-off between cost and validity for Algorithm~\ref{therecoursealgorithm} and the baselines.\label{fig:trade-off-app}}
\end{figure}

\end{document}